\documentclass[runningheads]{llncs}
\usepackage{graphicx}

\usepackage{tikz}
\usepackage{comment}
\usepackage{amsmath,amssymb} 
\usepackage{color}
\usepackage{import,booktabs}
\usepackage[export]{adjustbox}

\usepackage[pagebackref=true,breaklinks=true,letterpaper=false,colorlinks=true,
bookmarksopen=true,  
linkcolor=blue,pagebackref]{hyperref}

\usepackage{microtype}
\usepackage{graphicx}

\usepackage{xspace} 

\usepackage{amsmath,amssymb,amsfonts}
\usepackage{etoolbox}

\usepackage[thmmarks, amsmath, thref]{ntheorem}
\usepackage{bbm}
\usepackage{bm}

\usepackage[normalem]{ulem} 

\usepackage{times}
\usepackage{epsfig}
\usepackage{graphicx}
\usepackage{subcaption}

 \usepackage[linesnumbered,algoruled,noend,noline]{algorithm2e}

\usepackage{diagbox}

\usepackage{lipsum}  

\usepackage{threeparttable}
\usepackage[singlelinecheck=true,justification=centering]{caption}
\usepackage{makecell}
\usepackage{multirow}
\usepackage{color,soul}
\usepackage{xcolor}  

\definecolor{dimgray}{rgb}{0.35, 0.35, 0.35}
\usepackage{pifont}

\subimport{./}{macros}

\usepackage[accsupp]{axessibility}  

\usepackage{float}

\begin{document}
\pagestyle{headings}
\mainmatter
\def\ECCVSubNumber{4790}  

\title{A Deep Moving-camera Background Model}

\titlerunning{A Deep Moving-camera Background Model}
%
\author{Guy Erez\inst{1}\orcidID{0000-0002-4545-6664} \and
Ron Shapira Weber\inst{1}\orcidID{0000-0003-4579-0678} \and
Oren Freifeld\inst{1}\orcidID{0000-0001-9816-9709}}
\authorrunning{Erez et al.}
%
\institute{Ben-Gurion University of the Negev, Be'er Sheva, Israel \\
\email{\{ergu,ronsha\}@post.bgu.ac.il, orenfr@cs.bgu.ac.il}}
\maketitle

\begin{abstract}
In video analysis, background models have many applications such as background/foreground separation, change detection, anomaly detection, tracking, and more. However, while learning such a model in a video captured by a static camera is a fairly-solved task, in the case of a Moving-camera Background Model (MCBM), the success has been far more modest due to algorithmic and scalability challenges that arise due to the camera motion. Thus, existing MCBMs are limited in their scope and their supported  camera-motion types. These hurdles also impeded the employment, in this unsupervised task, of end-to-end solutions based on deep learning (DL). Moreover, existing MCBMs usually model the background either on the domain of a typically-large panoramic image or in an online fashion. Unfortunately, the former creates several problems, including poor scalability, while the latter prevents the recognition and leveraging of cases where the camera revisits previously-seen parts of the scene. This paper proposes a new method, called DeepMCBM, that eliminates all the aforementioned issues and achieves state-of-the-art results. Concretely, first we identify the difficulties associated with joint alignment of video frames in general and in a DL setting in particular. Next, we propose a new strategy for joint alignment that lets us use a spatial transformer net with neither a regularization nor any form of  specialized (and non-differentiable) initialization. Coupled with an autoencoder conditioned on unwarped robust central moments (obtained from the joint alignment), this yields an end-to-end regularization-free  MCBM that supports a broad range of camera motions and scales gracefully. We demonstrate DeepMCBM's utility on a variety of videos, including ones beyond the scope of other methods. 
Our code is available at \url{https://github.com/BGU-CS-VIL/DeepMCBM}. 

\end{abstract}

\begin{figure}[t]
\centering
\subcaptionbox{{Examples for several input frames}}[1\linewidth]
{
\includegraphics[width=0.3\linewidth,trim={0mm 0mm 0mm 0mm},clip]{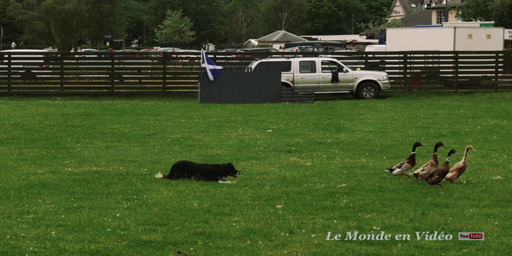}
\includegraphics[width=0.3\linewidth,trim={0mm 0mm 0mm 0mm},clip]{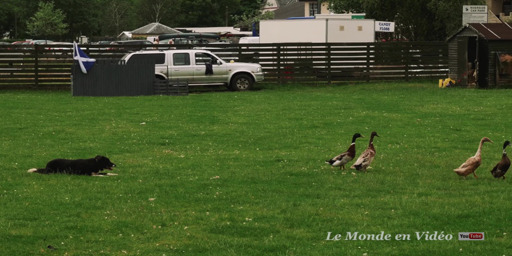}
\includegraphics[width=0.3\linewidth,trim={0mm 0mm 0mm 0mm},clip]{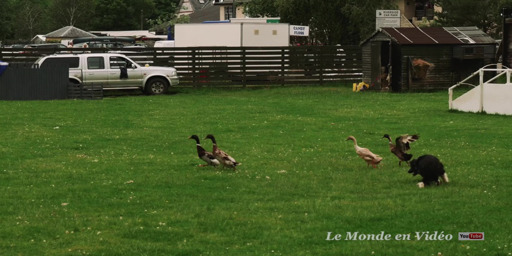}
}

\subcaptionbox{{Alignment, visualized via the mean panoramic image (computed from the entire video)}}[1\linewidth]
{
\includegraphics[width=0.90\linewidth,height =0.2\linewidth ,trim={0mm 0mm 0mm 0cm},clip]{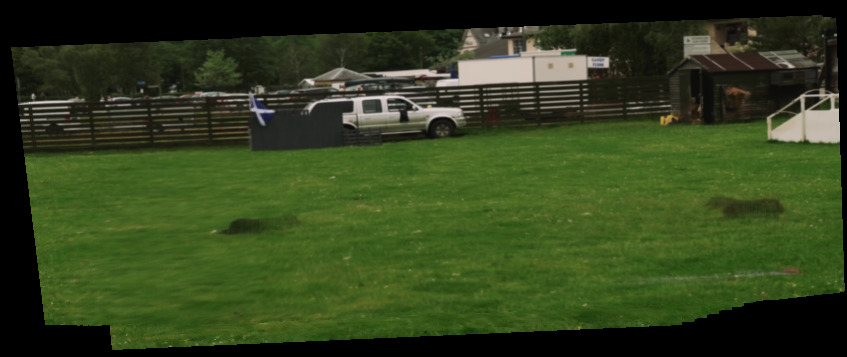}
}

\subcaptionbox{{Background estimation using the Conditional Autoencoder}}[1\linewidth]
{
\includegraphics[width=0.3\linewidth,trim={0mm 0mm 0mm 0mm},clip]{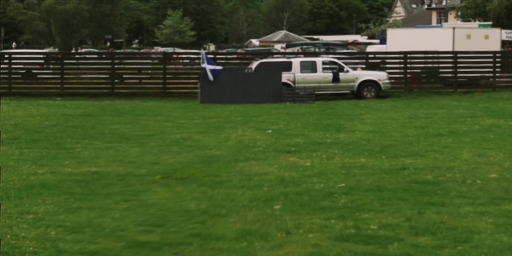}
\includegraphics[width=0.3\linewidth,trim={0mm 0mm 0mm 0mm},clip]{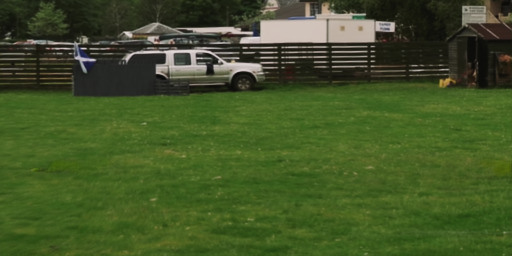}
\includegraphics[width=0.3\linewidth,trim={0mm 0mm 0mm 0mm},clip]{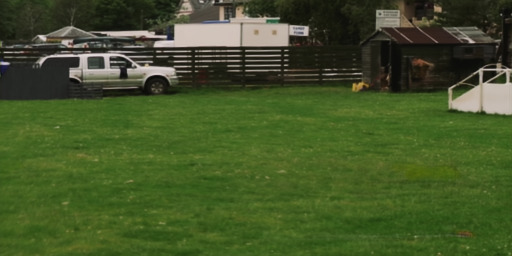}
}
\captionsetup{justification=centering, singlelinecheck=false}
\caption{Typical results of the proposed module. Note that despite the fact that the dog 
spent long times being static in two locations (as is evident by the corresponding ghosting effects in (b)) the model succeeded in eliminating it from the background.}
\label{Fig:Intro}
\end{figure}

\newcommand{\footnotestandalone}[1]{\let\thefootnote\relax\footnotetext{#1}}

\section{Introduction}\label{Sec:Intro}
The unsupervised video-analysis task this paper focuses on is learning 
 a background model in a video captured by a moving camera. 
In the simpler case where the camera is static,
such models have been used successfully  in many computer-vision applications
such as background/foreground separation, change or anomaly detection, and tracking.  
Static-camera solutions, however, cannot be easily extended to the moving-camera case 
since we do not know, a-priori, how the video frames
should be aligned to each other. Thus, most of the tools traditionally used in background models become less applicable; 
\eg, methods based on learning a low-dimensional subspace via Robust Principal Component Analysis (RPCA) assume that the 
 frames are aligned to each other.  

Seemingly, there is a straightforward solution: ``simply'' align the frames 
to each other to reduce the problem back to the static-camera case, and then build a static-camera background model based on the aligned frames. 
However, this is more complicated than it might seem. 
First, the alignment problem itself is often difficult. For example, methods based on creating a panoramic image by sequentially aligning each pair of consecutive frames 
suffer from drift errors. Moreover, such methods cannot exploit the information
conveyed in situations where the camera revisits (possibly from a different viewpoint)
a previously-seen region in the scene. 
This, among other considerations, motivates solutions based on Joint Alignment (JA) of the frames. However, even in this formulation the problem is often still hard to solve, partially due to reasons we analyze
later in~\autoref{Sec:JAisHard}. 
Second, and regardless of how the alignment is done, there is the issue of scalability
which pertains to not only the alignment problem itself but also the subsequent learning of the background model: when the accumulative motion of the camera throughout the video is substantial,
the domain of the panoramic image can be huge so background models 
learned in that domain must scale gracefully. Furthermore, in such cases, 
when a frame is warped (\ie, aligned) towards the panorama, it captures only a small portion of the latter. This means that \emph{most} of the data in the panoramic version
of the warped images is missing. This is problematic in our context since existing solutions for subspace learning in the presence of missing data usually struggle in such cases. 
Therefore, the missing-data issue, together with the scalability requirement, considerably complicates the task.  
Due to the above reasons, the success in the case of a \textbf{Moving-camera Background Model (MCBM)} is lagging far behind its static-camera counterpart.
Moreover, the difficulties above have also largely prevented the use of Deep Learning (DL) for this task. This is unfortunate not only because the idea of harnessing the power of DL 
is attractive but also since it hinders the usage of MCBMs within larger end-to-end pipelines. 

With this in mind, the goal of this paper is to provide an effective and scalable 
DL-based MCBM.  To that aim, we start by identifying more precisely what makes JA of video frames challenging: first in the general 
case and then in the more specific DL context. 
Next, we design a new JA strategy based on a regularization-free Spatial Transformer Net (STN) and a JA loss involving a memory aspect. Our method  requires no  
auxiliary tools (such as the brittle and non-differentiable initialization used in~\cite{Chelly:CVPR:2020:JA-POLS}) that would prohibit its usage 
within end-to-end pipelines. 
We also propose a new deep module for learning a background model. 
The model, based on a Conditional Autoencoder (CAE) and the output of the JA module, is learned in the small domain of the input frames instead of the much-larger panoramic domain. This eliminates scalability issues and targets the goal of estimating the background more directly.  
Importantly, this module too can be used within end-to-end pipelines. 
\autoref{Fig:Intro} demonstrates
the type of results obtained by the proposed modules.
Taken together, the proposed two modules give rise to a new and highly-effective MCBM method, coined \textbf{DeepMCBM},  which supports a broad range of camera motions and scales gracefully. We demonstrate DeepMCBM's utility on a variety of videos, including ones beyond the scope of competing methods.

\textbf{Our key contributions are:}
1) a DL module, for jointly aligning video frames, that relies on an STN-based optimization and a new training strategy that requires neither regularization nor initialization; 
2) a DL background-modeling module that leverages the JA via a CAE conditioned on unwarped robust central moments derived from the JA;  
3) together, these two modules form an end-to-end unsupervised MCBM that achieves SOTA results,
that scales gracefully, and that supports a wide range of camera motions. 

 \section{Related Work}\label{Sec:RelatedWork}
\textbf{STN}~\cite{Jaderberg:NIPS:2015:spatial} is a DL module that learns and applies a parameterized 
 input-dependent spatial transformation. 
Given a parameterized transformation family and an input image 
$f$, 
the STN's output consists of
a parameter vector $\btheta$ and a warped image obtained by warping $f$ using $T^\btheta$ (a transformation parameterized by $\btheta$). 
During training, the differentiation of a loss 
propagates through the STN.
In practice, however, and despite their elegance, potential strength, and usage in numerous papers, STNs are often hard to train.  
Part of our solution addresses exactly such a case, where we take an STN-based optimization problem that was thought to be too difficult~\cite{Chelly:CVPR:2020:JA-POLS} and show how it can, in fact, be solved easily, without resorting to a regularization or a sophisticated  limiting initialization. 

\textbf{Static-camera background models.} Early methods were pixelwise
(\eg, \cite{Stauffer:CVPR:1999:Adaptive}) but later the focus has shifted
to subspace estimation using 
Robust Principal Component Analysis and its variants (\eg,~\cite{DelaTorre:ICCV:2001:RPCA,Candes:JACM:2011:RPCA,Zhou:2010:rpca_candes_2,Guyon:ACCV:2012:rpca_low_rank}). 
While those models usually do not scale well, there also exist scalable RPCA models (\eg, \cite{Hauberg:CVPR:2014:TGA,Chakraborty:CVPR:2017:Intrinsic}). 
\\
\indent
\textbf{Image alignment.} In~\cite{Cuevas:ICCE:2015:statistical_object_detection,Moore:TCI:2019:PRPCA}, pairwise homographies are estimated between consecutive frames while~\cite{Jin:2008:multi_layer_homography} uses a multi-layer homography. 
An adaptive panoramic image is built in~\cite{Xue:2010:panoramic_bg_PTZ,Meneghetti:2015:panoramic_stitching2} while~\cite{Thurnhofer-Hemsi:2017:panoramic_ptz_competitive_net} relies on the assumption that a PTZ camera is used.
Most of the works above make stringent assumptions about the camera motion 
and estimate transformations between pairs of images, sometimes
even sequentially.  This approach, however, can lead to accumulative errors and/or significant distortions. 
To avoid such issues, AutoStitch ~\cite{Brown:IJCV:2007:AutoStitch} employs bundle adjustment. However, publicly-available implementations of AutoStitch  scale poorly with the number of images (\eg, cannot handle more than a few hundreds of frames). This is unlike the proposed approach which scales gracefully. 
Alignment methods relying on depth or expensive 3D information/reconstruction include~\cite{Newcombe:2011:SLAM,Klein:2007:SLAM2,Kendall:ICCV:2015:PoseNet,Wu:3DV:2013:SfM}. Unlike those works,
and similarly to, \eg, \cite{Chelly:CVPR:2020:JA-POLS}, the JA approach in this paper is purely 2D-based. 
\\
\indent
\textbf{MCBMs.} 
Online RPCA methods (\eg,~\cite{Balzano:Allerton:2010:GROUSE,He:CVPR:2012:GRASTA,Guo:2013:PRAC}) were extended to the case of camera jitter~\cite{He:2014:t-GRASTA}
as well as more significant motions~\cite{Gilman:ICCVw:2019:Panoramic}. 
DECOLOR~\cite{zhou:2012:DECOLOR} is another MCBM, based on motion detection, that is restricted to small motions.
IncPCP-PTI~\cite{Chau:2017:incPCP_PTI} targets
a PTZ-camera setting by updating a low-dimensional subspace with the help of an estimated
 rigid motion between consecutive frames. 
Several MCBMs are built by 
first aligning the frames to each other, 
and then, in the usually-large domain of the obtained panoramic image, 
learning a background model from the warped images
using a static-camera background model that can handle missing data (since each warped image covers only a portion of the panoramic domain). 
A prime example for such methods is PRPCA~\cite{Moore:TCI:2019:PRPCA}. 
Also of note are methods targeting \textbf{moving-object detection in a moving camera};
\eg, \cite{Yalcin:2005:flow_vehicle_detection,Sheikh:2009:object_detection_flow,Berger:2014:Subspace}. These works, however, cannot detect
changes unrelated to motion and also do not scale well. 
	\\
\indent
\textbf{STN-based JA.}
As we explain in~\autoref{Sec:JAisHard}, STN-based JA poses several difficulties. 
On that note, the closest work to ours is JA-POLS~\cite{Chelly:CVPR:2020:JA-POLS} which handles some of the difficulties via the usage of a non-differentiable and non-robust initialization, together with a fairly-restrictive regularization. While JA-POLS is effective in cases where it is applicable, it is limited in the camera-motion types it supports and is not an end-to-end solution. We will return to JA-POLS in more detail later on. 
\\
\indent
\textbf{Learning background models in the panoramic domain.}
Once alignment is obtained, in principle a background model can be learned. 
However, panoramic-size models (\eg,~\cite{Moore:TCI:2019:PRPCA}) do not scale
while using an ensemble of Partially-overlapping Local Subspaces (POLS)
\cite{Chelly:CVPR:2020:JA-POLS} is cumbersome
and also suffers from the fact the number of models grows with the size of the panorama. Either way, the existing methods do not offer an end-to-end solution that can be used easily within DL pipelines.

 \section{Preliminaries: Joint Alignment (JA)} 
Let $(f^n)_{n=1}^N$ be the frames of the input video and assume the size of each frame is $h\times w$ pixels. Let $C$ be the number of input channels; \eg, $C=3$  for RGB images (the case
considered in this paper).
Let $\Omega\subset\Rtwo$ denote the rectangular $h\times w$ common domain of each $f^n$, and let $\btheta_n$ denote the (latent) parameter vector 
of the spatial transformation associated with the sought-after alignment of  $f^n:\Omega\to\RR^C$.
The transformation itself, denoted by $T^{\btheta_n}$,
is viewed as an $\Rtwo\to\Rtwo$ map (not just $\Omega\to\Rtwo$).
The value of $d=\dim(\btheta_n)$ depends on the transformation family; \eg, in the affine case, $d=6$. 
The warped version of $\Omega$ is $\Omega^n\triangleq T^{\btheta_n}(\Omega)\triangleq\set{\bx:\exists \bx'\in \Omega \text{ s.t.~}T^{\btheta_n}(\bx')=\bx}\subset \Rtwo$.
Mathematically, we define the warped image
as $g^n:\Omega^n \to \RR^C$ using the equality
\begin{align}
 g^n(T^{\btheta_n}(\bx')) = f^n(\bx') \quad \forall \bx'\in\Omega \, . 
\end{align}
However, due to technical reasons related to image warping~\cite{Szeliski:Book:2010}, 
it is more convenient and customary  to define $g^n$ via the inverse transformation of $T^{\btheta_n}$: 
\begin{align}
g^n_\bx\triangleq
 g^n(\bx) = f^n((T^{\btheta_n})^{-1}(\bx)) \quad \forall \bx\in\Omega^n \, . 
\end{align}
Note that $g^n$ depends on $\btheta_n$ and $f^n$. 
Let $H$ and $W$ be the height and width, respectively, of a rectangle,
denoted by $\Omega_\mathrm{scene}\subset \Rtwo$,  that is large enough to contain $\bigcup_{n}\Omega^n$. 
We now define a mask that will be useful
for reasons to become clear shortly. 
Let $M^\Omega$ be a single-channel $h\times w$ image whose domain is $\Omega$ and whose values are all equal to 1.  
Let $M^n:\Omega_\mathrm{scene}\mapsto[0,1]$ be a non-binary $H\times W$   mask
 obtained by image warping of $M^\Omega$, according to $T^{\btheta_n}$, using zero padding and a bilinear interpolation kernel. That is, 
 for any integral location $\bx$ in $\Omega_\mathrm{scene}$, the value of $M^n$ at $\bx$, denoted by $M^n_\bx$, is given by
 \begin{align}
 M^n_{\bx}=\widetilde{M^\Omega}_{\bx'} \qquad \bx'=T^{-\btheta_n}(\bx)\in\Rtwo\, \end{align}
 where  $\widetilde{M^\Omega}_{\bx'}$ 
 is interpolated from the values of  $M^\Omega$ at the 4 integral locations nearest to $\bx'$ where whenever any of those integral locations
 falls outside $\Omega$ 
 the value of $M^\Omega$ at that location is taken to be zero.  
 Thus, $ M^n_{\bx}= 0$ if all those 4 locations are outside $\Omega$,
 $M^n_{\bx}=1$ it they all fall inside it, and $0<M^n_{\bx}< 1$ otherwise. 
Let $g^n_{\bx,c}$ denote the value of $g^n_\bx$ at channel $c$. 
We will refer to 
$
p_{\bx,c}\triangleq (g^n_{\bx,c})_{n=1}^N$
where $
c\in \set{1,\ldots,C}
$
as the $C$ pixel stacks at location $\bx$. 
Similarly, we define the mask stack at location $\bx$ as
$
  m_{\bx}\triangleq (M^n_{\bx})_{n=1}^N\, .
$
    %
Note that $p_{\bx,c}$ and $m_\bx$ depend on 
$(\btheta_n)_{n=1}^N$. 
A \textbf{joint-alignment loss}, to be minimized \wrt $(\btheta_n)_{n=1}^N$,  may be formulated in terms of
\begin{align}
\Lcal_{JA}=
  \mathrm{func}( ((p_{\bx,c})_{c=1}^C,m_{\bx})_{\bx\in \Omega_\mathrm{scene}})
\label{Eqn:JA_loss_in_the_general_case}\, .
  \end{align}
 For example, in the early works on \emph{congealing} (\eg,~\cite{Miller:CVPR:2000:LearningOneExample,Learned:PAMI:2006:DataDrivanViaJA,Huang:CVPR:2007:unsupervised,Huang:NIPS:2012:learning}) 
 that loss was based on entropy minimization. Later, other researchers~\cite{Cox:CVPR:2008:LS,Cox:ICCV:2009:LS}
 showed the benefits of a loss based on least squares.  
 A robust variant (used in~\cite{Chelly:CVPR:2020:JA-POLS}) of the latter  is
\begin{align}
\Lcal_{JA}=
  \frac{1}{N}\sum\nolimits_{n=1}^N
  \frac{1}{C}
  \sum\nolimits_{c=1}^C 
  \frac{\sum_{\bx\in \Omega_\mathrm{scene}} M^n_{\bx}
  \rho_\mathrm{JA}( g^n_{\bx,c}-\mu_{\bx,c})}{\sum_{\bx\in \Omega_\mathrm{scene}}M^n_{\bx}}
  \label{Eqn:JALossFullData}
\end{align}
where
$\mu_{\bx,c} = \frac{\sum_{n=1}^{N} M^n_{\bx}g^n_{\bx,c}}
{\sum_{n=1}^{N} M^n_{\bx}}$
and $\rho_\mathrm{JA}$ is a differentiable robust error function~\cite{Black:IJCV:1996:Robust}.

Let $\mu$ be the mean of the warped images; \ie, the  value
of $\mu$ at location $\bx$ and channel $c$
is $\mu_{\bx,c}$. Note that $\mu$ may be viewed as the ``moving target''
to which the frames should be aligned. It ``moves'', 
during the optimization, in the following sense. 
As the alignment of the frames keeps changing, $\mu$ changes too since it is computed using the (weighted) average of the warped images. 
Assuming that the parameterization $\btheta_n\to T^{\btheta_n}$ is differentiable and
that the transformation family is sufficiently well-behaved (as is the case, \eg, with the affine group or, more generally, spaces of diffeomorphisms~\cite{Freifeld:ICCV:2015:CPAB,Freifeld:PAMI:2017:CPAB,Skafte:CVPR:2018:DDTN,Shapira:NIPS:2019:DTAN,Kaufman:ICIP:2021:cyclic}), the loss in~\EQN\eqref{Eqn:JALossFullData} is differentiable. 
Thus, if $\btheta_n$ is predicted using an STN
(so, in particular, $\btheta_n$ is a differentiable function of $f^n$, the STN's input),
the loss can, at least in principle, 
be minimized using standard DL training.

 \section{Identifying Key Challenges in Solving Joint-alignment Problems}\label{Sec:JAisHard} 
Below we discuss three issues that might arise when solving JA problems:
\textbf{1)} poor \emph{global} minima; 
\textbf{2)} the need of regularization;
\textbf{3)} the need  of a good initialization. 
\begin{figure}[t]
\newcommand{\Size}{0.19\linewidth}
\centering
\subcaptionbox{A typical problem: if $\Omega_\mathrm{scene}$ is not very large,
the process is prone to a poor \emph{global} minimum.\label{Fig:JAproblems:PoorGlobalMin}}[1\linewidth]
{
\includegraphics[width=\Size,trim={0mm 0mm 0mm 0mm},clip]{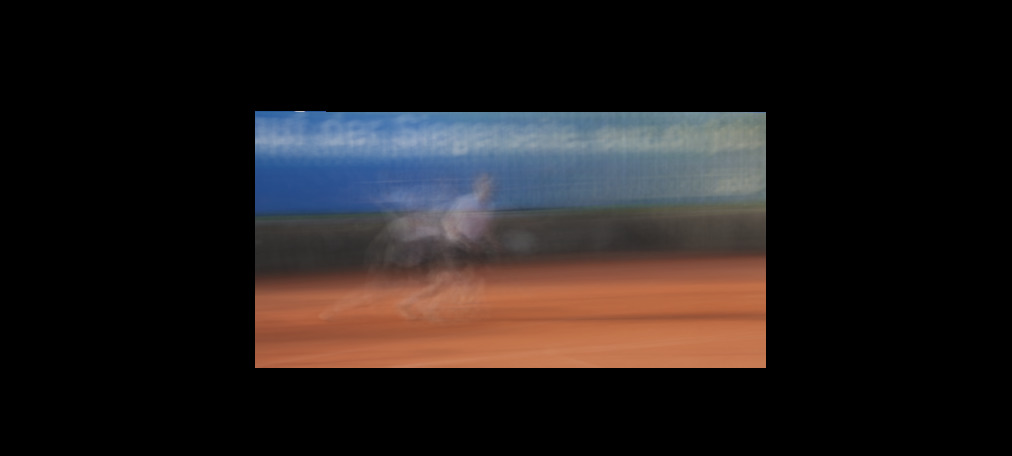}
\includegraphics[width=\Size,trim={0mm 0mm 0mm 0mm},clip]{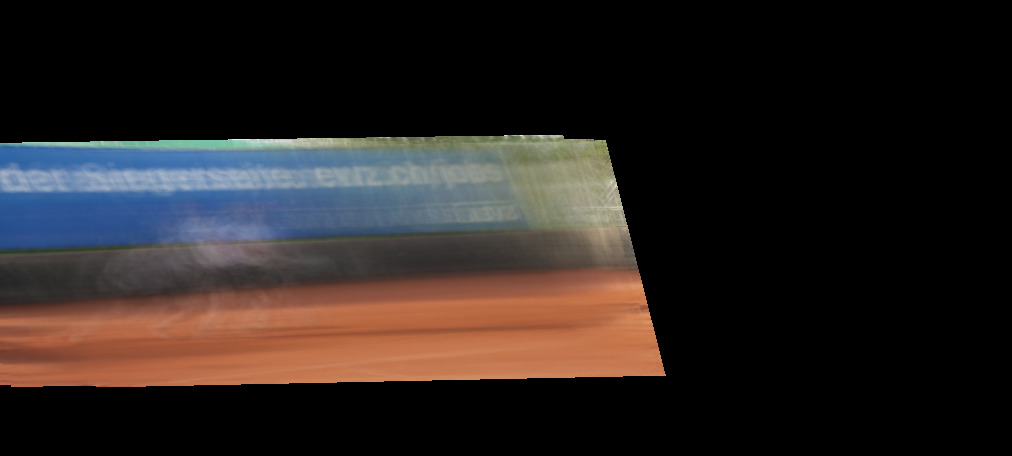}
\includegraphics[width=\Size,trim={0mm 0mm 0mm 0mm},clip]{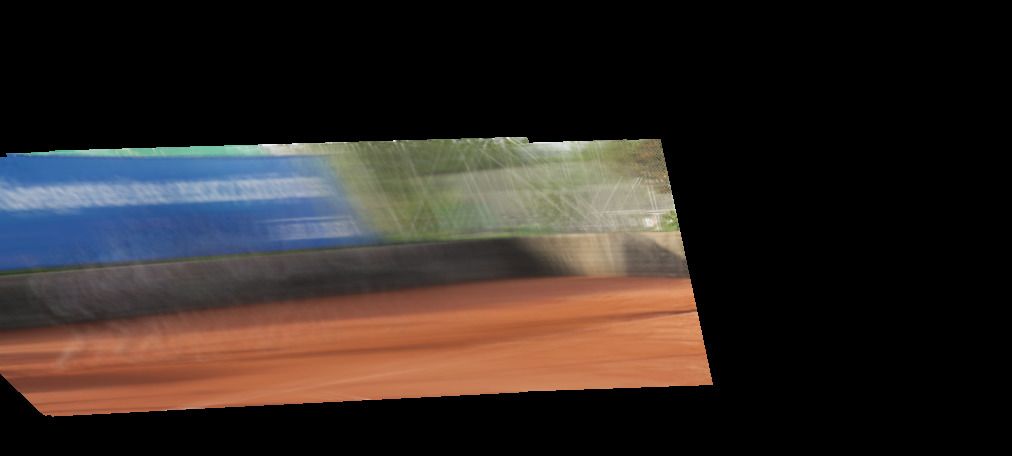}
\includegraphics[width=\Size,trim={0mm 0mm 0mm 0mm},clip]{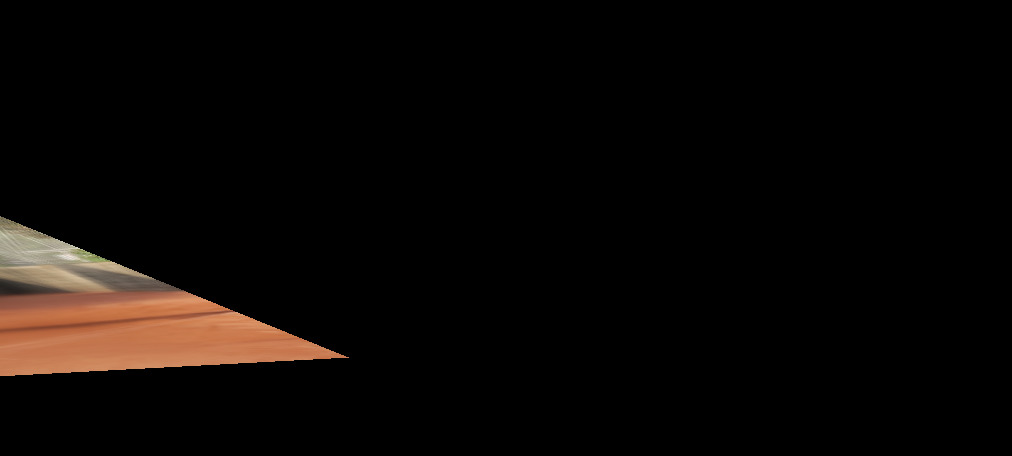}
\includegraphics[width=\Size,trim={0mm 0mm 0mm 0mm},clip]{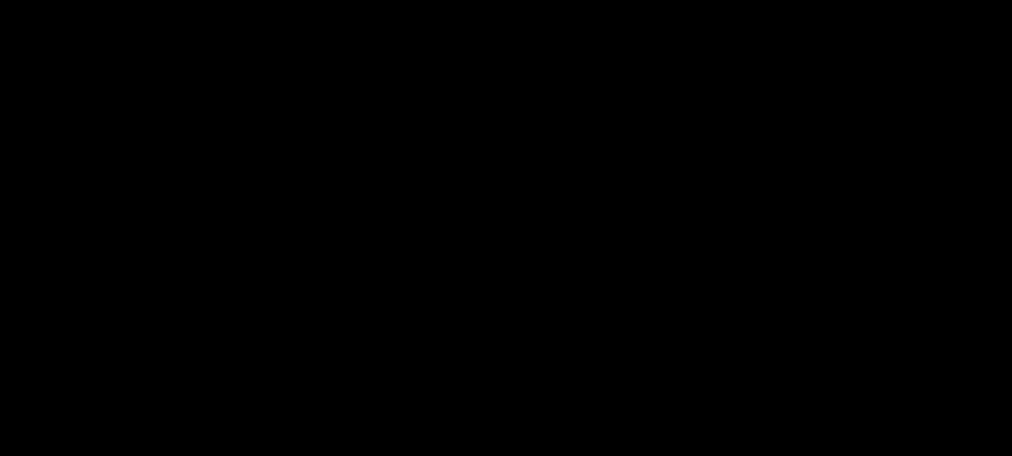}
}
\subcaptionbox{A typical problem: drastic spatial changes in $\mu$
(note also that the end result is quite blurry).
\label{Fig:JAproblems:Jumps}}[1\linewidth]
{
\includegraphics[width=\Size,trim={0mm 0mm 0mm 0mm},clip]{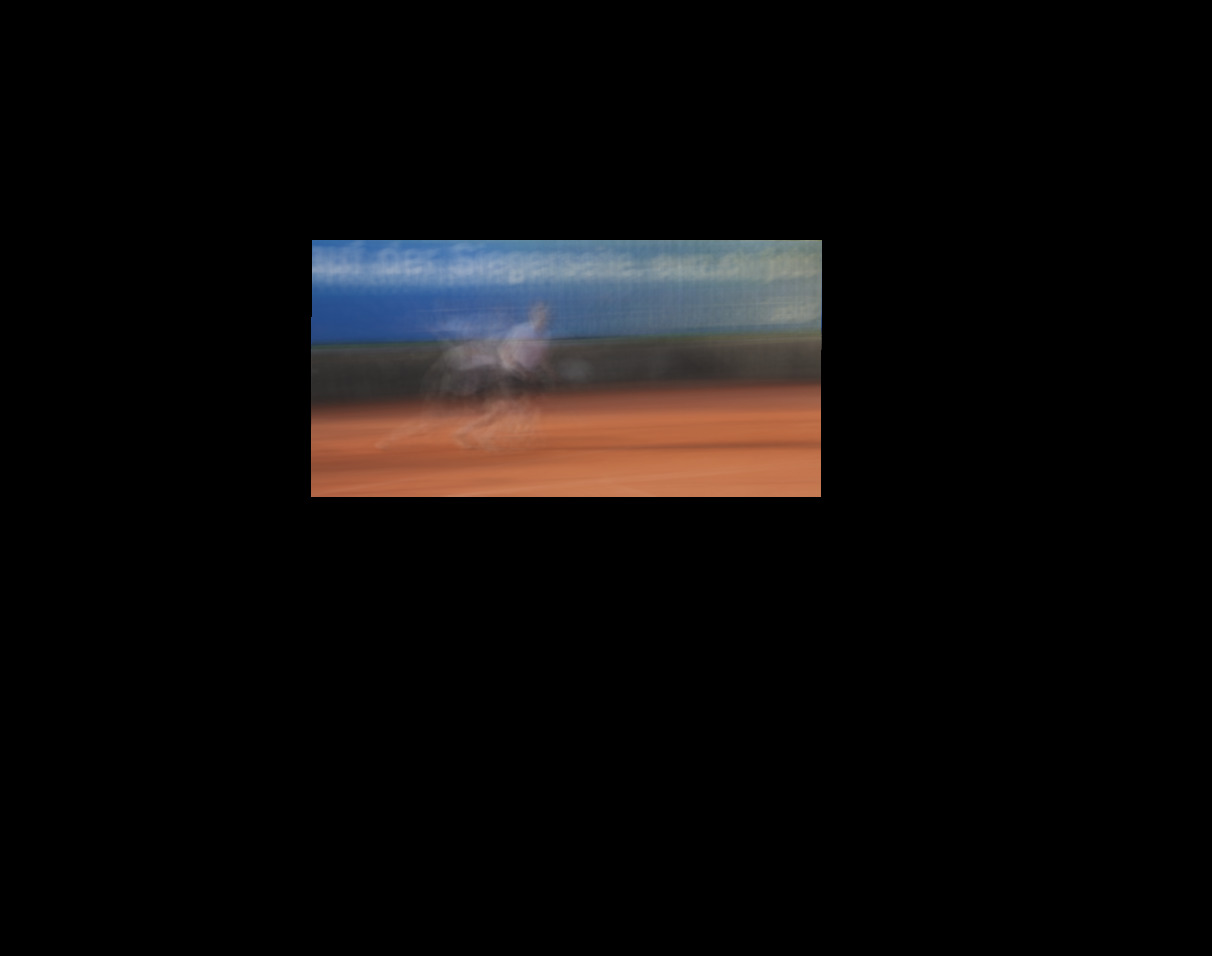}
\includegraphics[width=\Size,trim={0mm 0mm 0mm 0mm},clip]{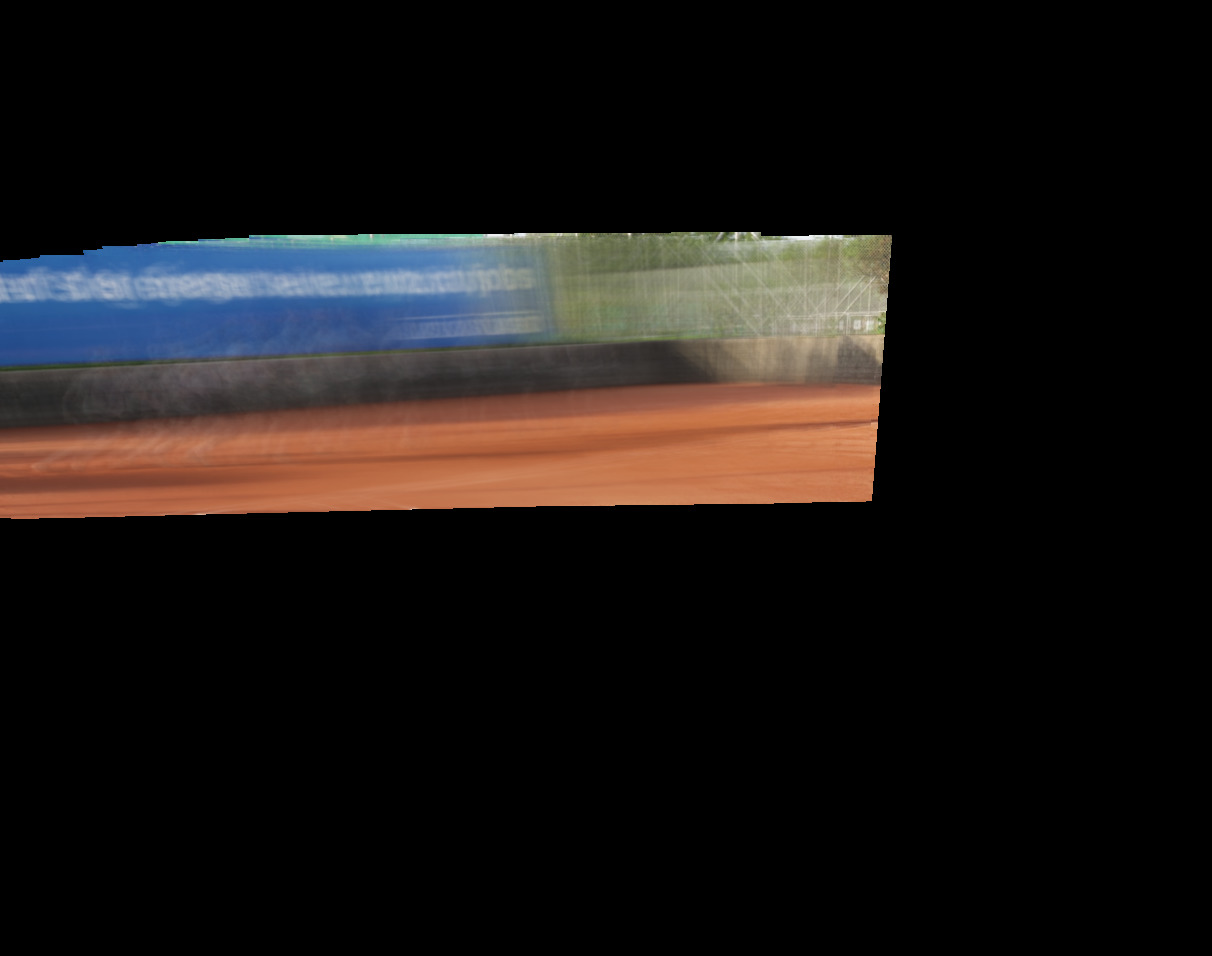}
\includegraphics[width=\Size,trim={0mm 0mm 0mm 0mm},clip]{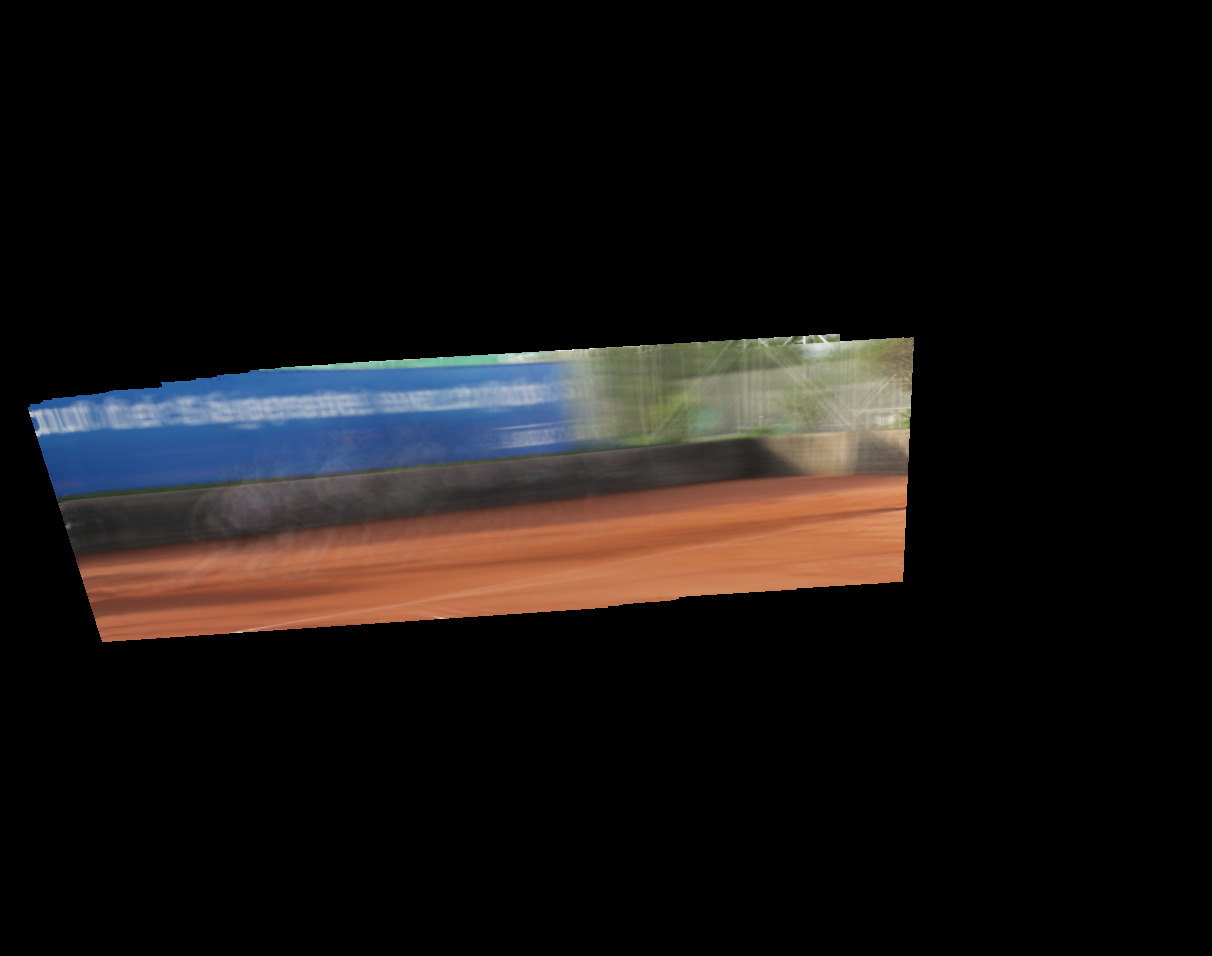}
\includegraphics[width=\Size,trim={0mm 0mm 0mm 0mm},clip]{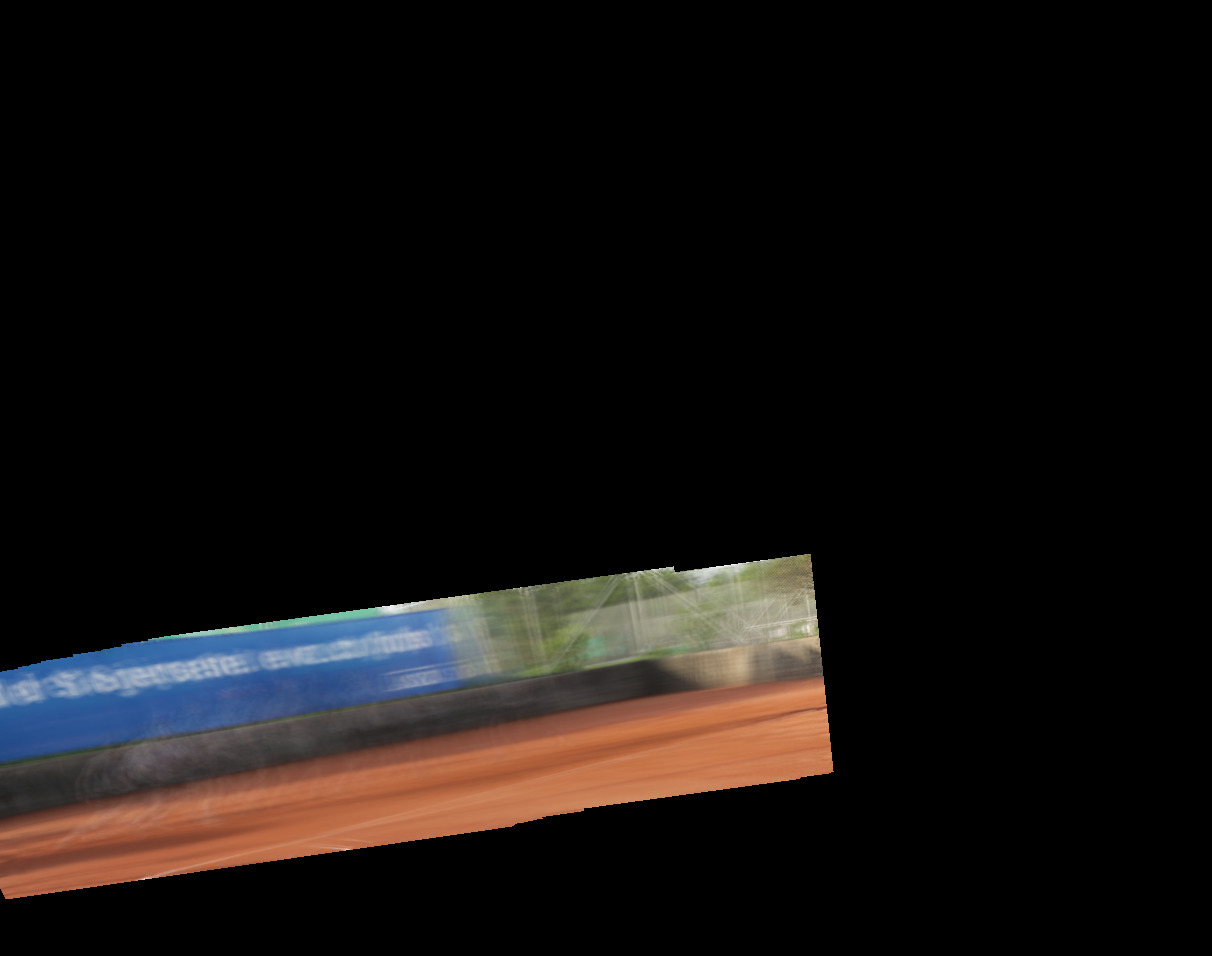}
\includegraphics[width=\Size,trim={0mm 0mm 0mm 0mm},clip]{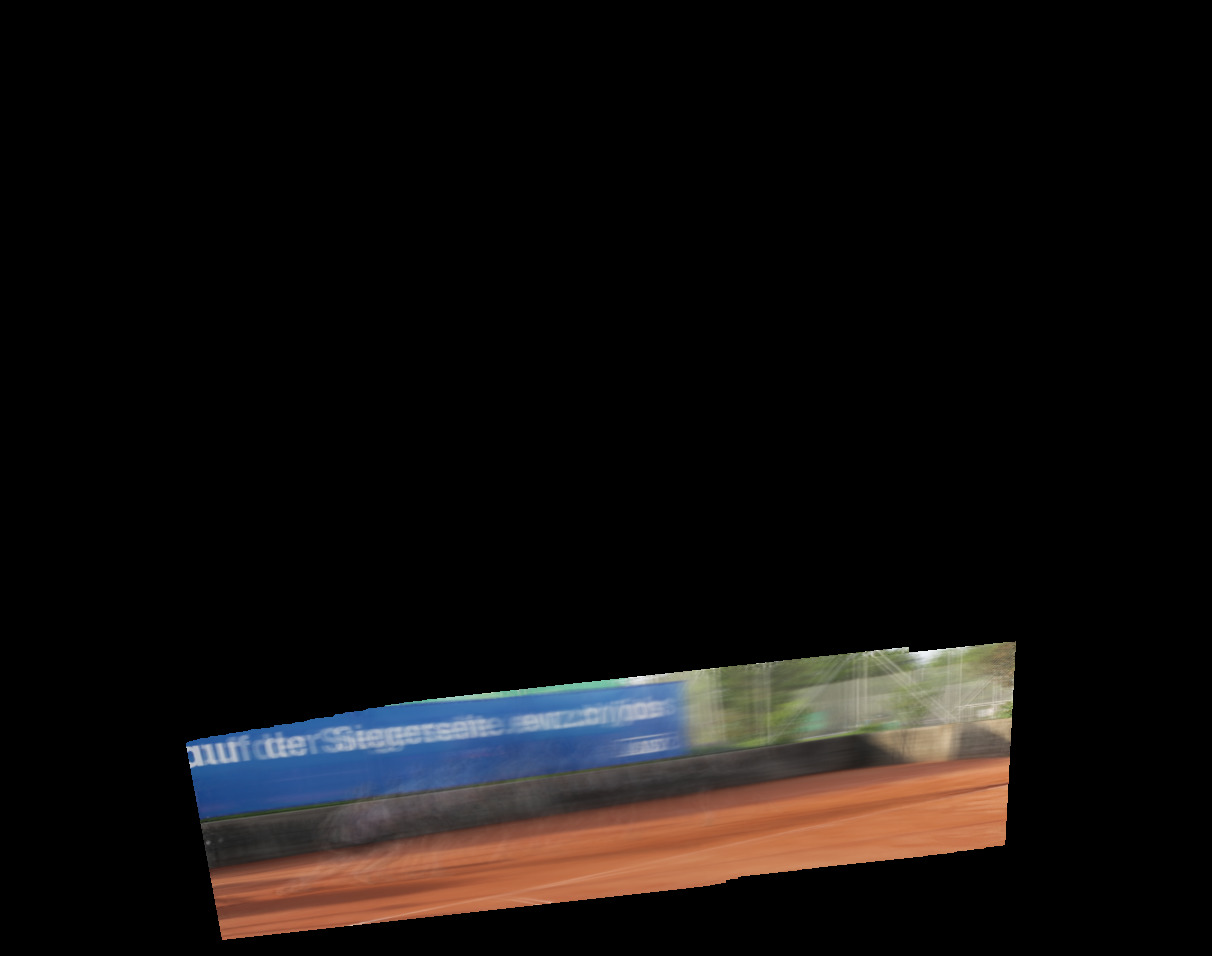}
}
\caption{Typical problems in JA. Rightmost images are post-convergence results.}
\label{Fig:JAproblems}
\end{figure}

Usually when trying to minimize a  loss, reaching a global minimum is hard or even impossible, and if this feat happens to be achieved, it is deemed to be the ultimate success. Sadly, global minima of $\Lcal_\mathrm{JA}$, while being (very) easy to achieve,  
reflect, in fact, an ultimate failure; \eg, the non-negative  $\Lcal_\mathrm{JA}$
can attain its global minimum (\ie, zero) when all the frames are shrunk to an infinitesimally-small point. 
A similar phenomenon occurs if all the frames are warped outside $\Omega_\mathrm{scene}$ (\eg, see~\autoref{Fig:JAproblems:PoorGlobalMin})  or if the frames are warped such that there will be no pairwise overlap between them. 

A popular solution in such cases is adding some type of regularization over $(\btheta_n)_{n=1}^N$. However, while various forms of regularization have been suggested, each of them imposes a certain bias; \eg, the regularization term in~\cite{Learned:PAMI:2006:DataDrivanViaJA} favors symmetric distributions 
while the one in~\cite{Chelly:CVPR:2020:JA-POLS} pushes the (affine) transformations
towards the Special Euclidean group, denoted by \SEtwo. The implied assumptions in both these cases are limiting. 
Likewise, penalizing the size of the transformations (\eg, by penalizing some norm
of $\btheta_n$) is problematic when the accumulative motion of the camera is large, while regularization favoring temporal smoothness 
is not always compatible with real camera motions.  
Another issue is the need of hyperparameter tuning for the weight of the regularization term.  
Moreover, finding a combination of a regularization type and a weight 
that will work well for a sufficiently-large variety of videos is  difficult. 

JA is usually a difficult non-convex problem. Thus, a good initialization can be  useful; \eg, in JA-POLS~\cite{Chelly:CVPR:2020:JA-POLS} 
an STN-based JA module had to rely on an initialization based on SE-Sync~\cite{Rosen:IJRS:2019:SESync}. 
The latter provides a useful globally-optimal solution to a different-but-related problem: the estimation of \emph{absolute} transformations that are consistent as possible
with noisy measurements of pairwise \emph{relative} transformations between pairs of frames, where both the latent absolute transformations and the observed relative ones are in \SEtwo. With that initialization, the STN needs to solve an easier problem and does so over the more expressive Affine group. 

There are, however, several  problems with the JA  approach
in~\cite{Chelly:CVPR:2020:JA-POLS} (we will later also discuss problems related to the background-modeling approach in~\cite{Chelly:CVPR:2020:JA-POLS}). 
First, pre-processing and heuristics are needed for extracting the relative transformations. 
Second, in cases where some of the true latent absolute transformations are far from \SEtwo\ (\eg: when the video contains a significant accumulative variation in the distance between the camera and the scene; when the camera zoom is changing;
when there is a strong perspective effect; \etc), the initialization breaks and this leads in turn to JA-POLS' failure. Moreover, SE-Sync is neither robust nor differentiable \wrt the input frames. As there is no easy way to differentiate SE-Sync \wrt the input frames, the STN-based JA module in JA-POLS cannot be used in an end-to-end DL pipeline. 
\subsection{An Additional Challenge with Joint Alignment When Using Batches} \label{Sec:JAinBatchesIsHard}
Typically, due to the data size and as it is almost always the case in DL, 
the optimization is done batch by batch 
where each batch consists of a subset (selected at random) of the frames from the entire video. A single epoch then represents a full pass over the entire data,  
and the frames are reshuffled between epochs.
This typically-necessary batch-by-batch processing creates an optimization difficulty which might appear 
to be minor but is, in fact, far more critical than it may seem (we will revisit this point in~\autoref{Sec:Method:OurJA}).   
The issue is that the mean image $\mu$ (from~\EQN\eqref{Eqn:JALossFullData}) is a function 
of the entire video, not just the frames in the current batch.
A seemingly-obvious solution is to hold $\mu$ fixed during each epoch -- so it does not affect the computation of the loss' gradient -- and then, at the end of each epoch, recompute $\mu$. 
However, a problem that arises with that approach is that the difference 
between the alignment targets (that is, the previous $\mu$ and the recomputed one) in each pair of consecutive epochs might be large, making the optimization difficult since the optimal transformations for one target might be quite far
from those that are optimal for the next target. For an illustration, see~\autoref{Fig:JAproblems:Jumps}.
A different approach, used in~\cite{Chelly:CVPR:2020:JA-POLS}, 
picks the target $\mu$ to be the mean of only the (warped) frames in the current batch. 
Besides the fact that this is somewhat inconsistent with the cost-function formulation, that approach too can cause significant changes in the targets between consecutive batches. 
The jumping-target problem complicates the optimization more than one may expect. This is especially an issue at the beginning of the  process when the frames are completely misaligned. For example,
in retrospect, this is partly why JA-POLS~\cite{Chelly:CVPR:2020:JA-POLS} had to rely on the SE-Sync-based initialization scheme: as shown in~\cite{Chelly:CVPR:2020:JA-POLS}, except in the simple case where the accumulative camera motion is small, without that initialization JA-POLS usually fails.

\section{The Proposed Method: DeepMCBM}\label{Sec:Method}
The proposed modules of 
joint alignment (using an STN) and background modeling (using a CAE) are presented in~\autoref{Sec:Method:OurJA} and~\autoref{Sec:Method:CAE}, respectively. 
Together, they form the proposed method, DeepMCBM. 
The goal of the STN straining is 1) to jointly align the video frames, implicitly forming a panoramic image,  and 2) to learn how to warp an input frame towards 
that panoramic image. 
The goal of the CAE training is to learn the variability in the differences between the panoramic image and the input frames, while taking the warping into account
but ignoring the foreground objects. The conditioning 
is done using the robust version of the panoramic pixelwise mean and variance.  

{    \SetKwComment{Comment}{}{}
    \begin{algorithm}[t]
    \KwIn{$N_\mathrm{epochs}$, $N_\mathrm{batches}$,   $\rho(\cdot)$, data\_loader}
    \KwData{$(f^n)_{n=1}^N$}
    \KwOut{A trained STN for Joint Alignment}
    Initialize accumulators $\Gcal\in \RR^{H\times W \times C}$ and $\Mcal\in \RR^{H\times W}$ $\quad$ \Comment{// see text}
    \For{$e\in \set{1,\ldots,N_\mathrm{epochs}}$}    {
        \For{$i\in \set{1,\ldots,N_\mathrm{batches}}$}    { 
             $(f^b)_{b\in B} \gets$  data\_loader $\quad$ \Comment{// Load batch: $B\subset\set{1,\ldots,N}$}
$(\btheta_b$, $g^b)_{b\in B} \gets \mathrm{STN}((f^b)_{b\in B})$  $\quad$ \Comment{// Note that $g^b=f^b \circ T^{\btheta_b}$}
     $(M^b)_{b\in B} \gets (M^\Omega \circ T^{\btheta_b})_{b\in B}$ 
     $\quad$ \Comment{// Warp masks}
        $\Gcal$, $\Mcal$,  $\Lcal_\mathrm{batch}\gets$ \autoref{Alg:create_mu_compute_batch_loss_update_accum}(
         $\Gcal$, $\Mcal$, $(g^{b})_{b\in B}$, $(M^{b})_{b\in B}$)
         \Comment{// Update $\Gcal$ and $\Mcal$; measure $\Lcal_\mathrm{batch}$ (\ie, the batch loss)}
         Perform an optimization step  to minimize the $\Lcal_\mathrm{batch}$ loss. 
    }
 $(\Gcal,\Mcal) \gets (\lambda\Gcal,\lambda\Mcal)$ $\quad$\Comment{// 
 Keep the history, but downweight it}\label{Alg:Update_target_and_compute_batch_loss:LineDownWeight}
}
\caption{Training an STN for Joint Alignment}   \label{Alg:Update_target_and_compute_batch_loss}
    \end{algorithm}
}

\subsection{A Regularization-free Strategy for Joint Alignment}\label{Sec:Method:OurJA}
Having identified, in~\autoref{Sec:JAinBatchesIsHard}, that the jumps in the values of $\mu$
cause a major difficulty in the STN-based optimization of
$\Lcal_\mathrm{JA}$  (\EQN\eqref{Eqn:JALossFullData}), we design a simple but surprisingly-effective optimization strategy, summarized 
in~\autoref{Alg:Update_target_and_compute_batch_loss} (which, in turn, uses
\autoref{Alg:create_mu_compute_batch_loss_update_accum} as its subroutine). 
During the training epochs, instead of computing $\mu$ using only the current batch (as was done in~\cite{Chelly:CVPR:2020:JA-POLS}), or instead of recomputing $\mu$ from scratch each epoch, we construct our $\mu$ from the warped frames in the current batch
while also taking into account, albeit with a lower weight, all the warped frames from the previous epochs as well as the previous batches in the current epoch.
The proposed algorithm uses accumulators, denoted by $\Gcal$ and $\Mcal$. 
The former is used to accumulate weighted sums of the values of the pixels in the warped 
frames while the latter serves a similar purpose
with the values of the pixels in the warped masks. 
Concretely, let $e$ denote the index of the current epoch 
and let $e'$ denote the index of some previous epoch. 
When evaluating the loss in a batch during epoch $e$, the contribution of the results from epoch $e'$ becomes smaller and smaller as the ``time'' difference, $e-e'$, grows.
This is done in~\autoref{Alg:Update_target_and_compute_batch_loss:LineDownWeight} 
in~\autoref{Alg:Update_target_and_compute_batch_loss}
by multiplying the accumulators of the warped frames and the warped masks by a positive factor $\lambda$ where $\lambda<1$ (we use $\lambda=0.9$). 
{    \SetKwComment{Comment}{}{}
    \begin{algorithm}[t]
    \KwIn{$\Gcal$, $\Mcal$, $(g^{b})_{b\in B}$, $(M^{b})_{b\in B}$}
    \KwOut{$\Gcal$, $\Mcal$,  $\Lcal_\mathrm{batch}$}
      $\Gcal \gets \Gcal + \sum_{b=1}^B g^{b}$ $\quad$ \Comment{// update warped-image accumulator}
   $\Mcal \gets \Mcal +  \sum_{b=1}^B M^{b}$ $\quad$ \Comment{// update warped-mask accumulator}
   $\mu \gets \bzero_{H\times W\times C}$ $\quad$ \\
    \ForPar{$\bx \in
    \set{\bx: \bx \in \Omega_\mathrm{scene} \text{ and } \Mcal_\bx \ge 0}    
    $}{
      \ForPar{$c \in \set{1,\ldots,C}$}{
       $\mu_{\bx,c}\gets \frac{\Gcal_{\bx,c}}{\Mcal_{\bx}}$
       }
    }
    $\Lcal_\mathrm{batch}\gets \frac{1}{B}\sum_{b=1}^B
  \frac{1}{C}
  \sum_{c=1}^C \left[
  \left(\sum_{\bx\in \Omega_\mathrm{scene}} M^b_{\bx}\rho( g^b_{\bx,c}-\mu_{\bx,c})\right)/\left(\sum_{\bx\in \Omega_\mathrm{scene}}M^b_{\bx}\right)\right]$\\

   \caption{Update ($\mu, \Gcal, \Mcal)$ and measure the loss on the batch}  
   \label{Alg:create_mu_compute_batch_loss_update_accum}
    \end{algorithm}
}

\begin{figure}[!t]
\newcommand{\Size}{0.19\linewidth}
\centering
\subcaptionbox{Compared with~\autoref{Fig:JAproblems:PoorGlobalMin}, the process is more stable and successful. Also, even when $\mu$ nears the border of $\Omega_\mathrm{scene}$, 
it never goes outside it. }[1\linewidth]
{
    \includegraphics[width=\Size,trim={0mm 0mm 0mm 0mm},clip]{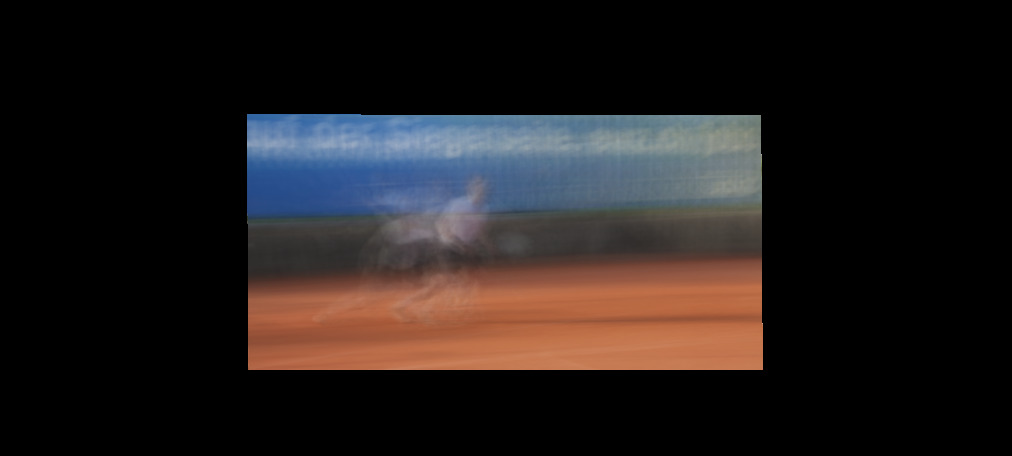}
    \includegraphics[width=\Size,trim={0mm 0mm 0mm 0mm},clip]{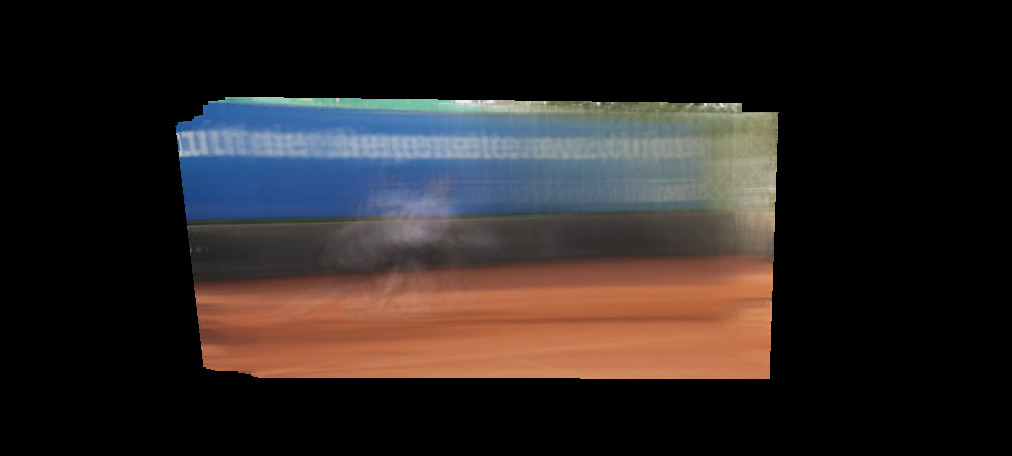}
    \includegraphics[width=\Size,trim={0mm 0mm 0mm 0mm},clip]{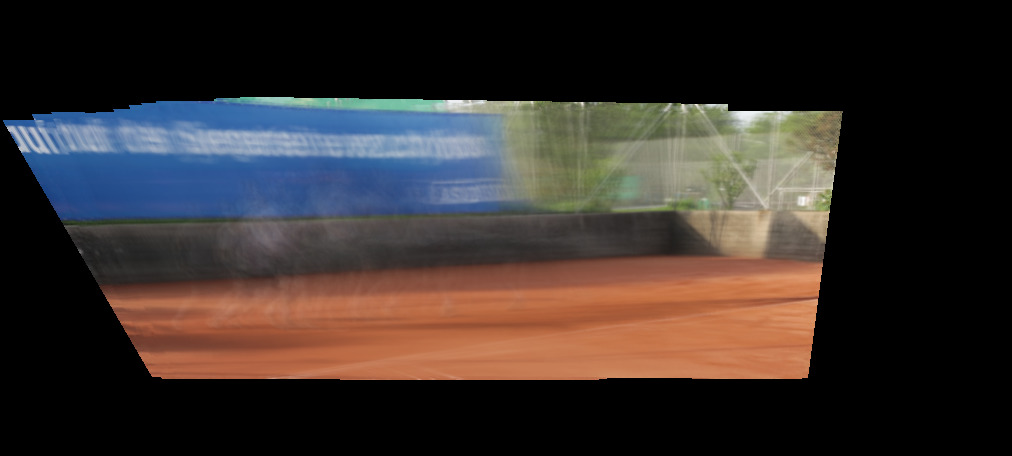}
    \includegraphics[width=\Size,trim={0mm 0mm 0mm 0mm},clip]{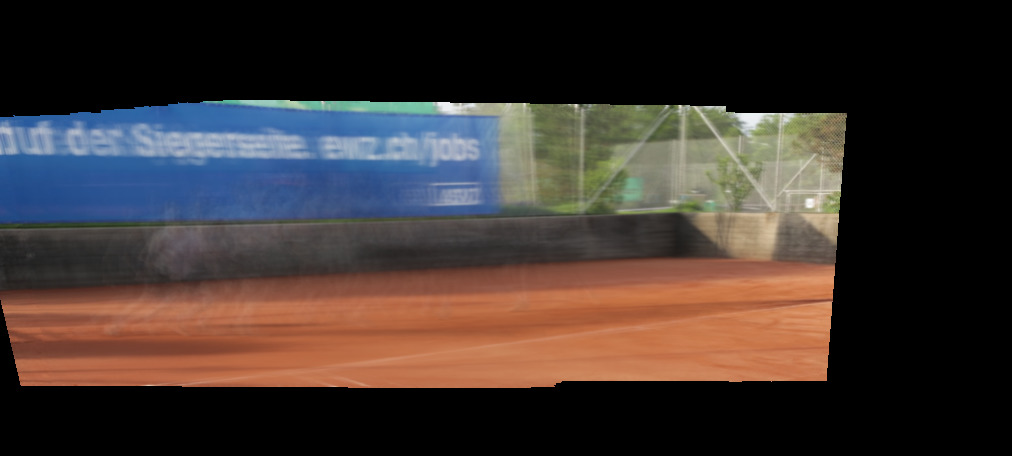}
    \includegraphics[width=\Size,trim={0mm 0mm 0mm 0mm},clip]{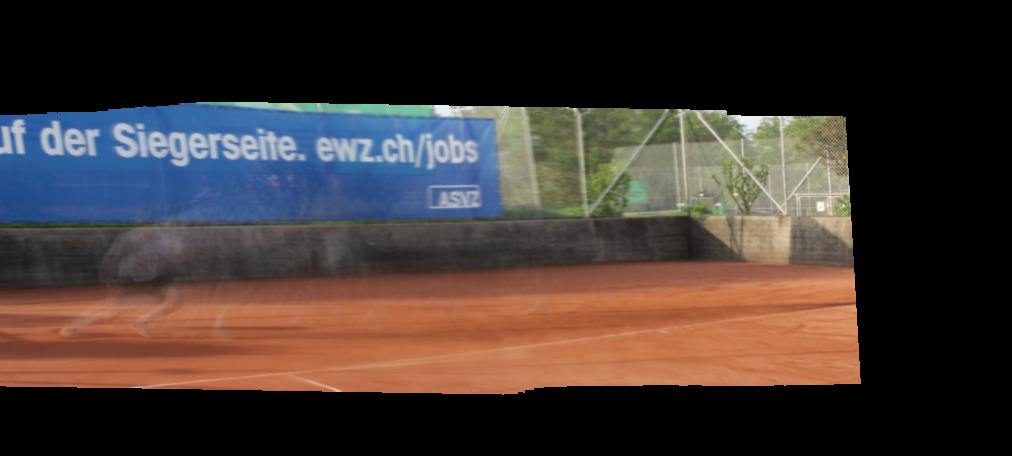}   
}
\captionsetup{justification=raggedright, singlelinecheck=false}
\subcaptionbox{Compared with~\autoref{Fig:JAproblems:Jumps}, the drastic jumps are eliminated. Also, with the proposed term the results are less affected
by the specified size of $\Omega_\mathrm{scene}$. }[1\linewidth]
{
    \includegraphics[width=\Size,trim={0mm 0mm 0mm 0mm},clip]{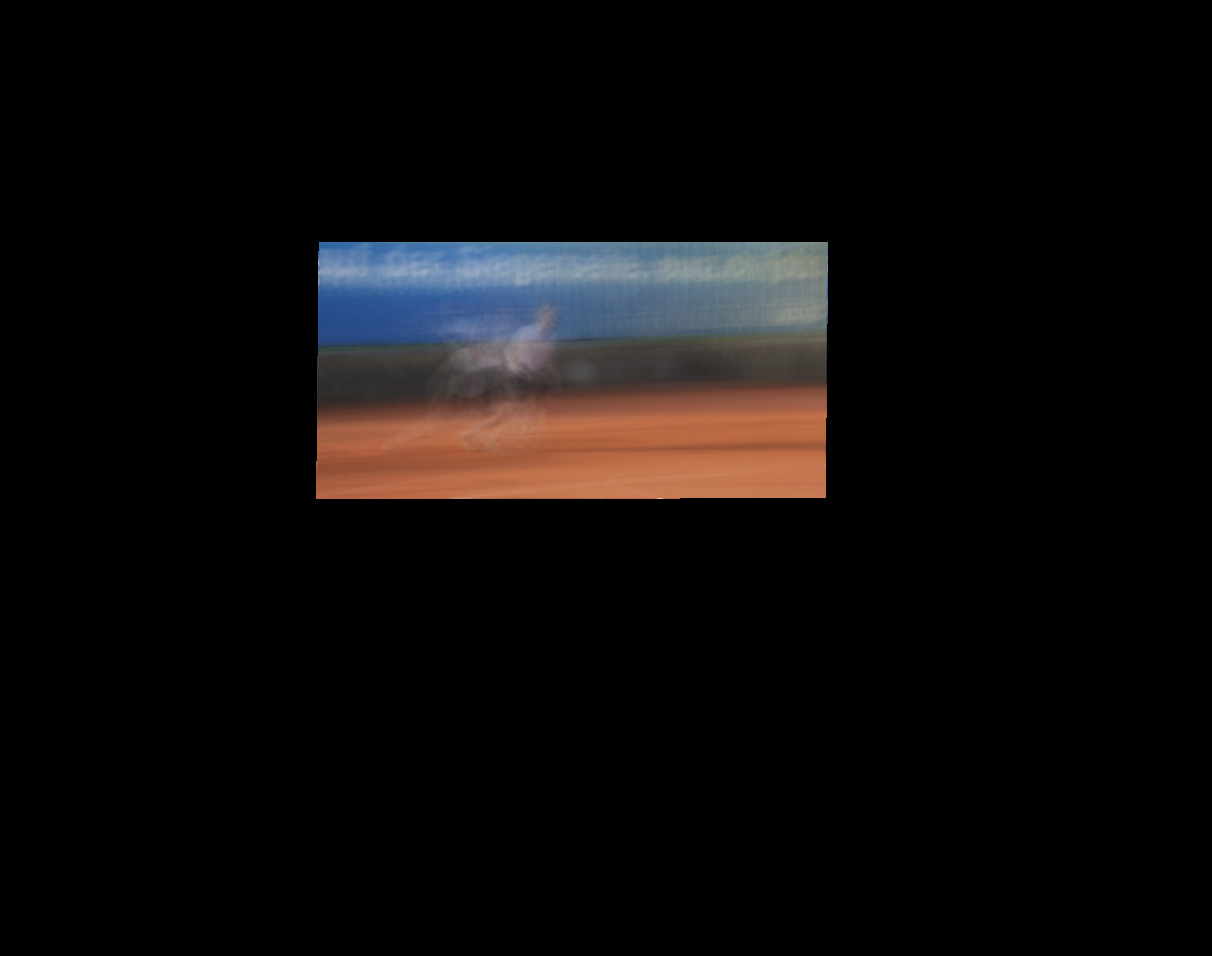}
    \includegraphics[width=\Size,trim={0mm 0mm 0mm 0mm},clip]{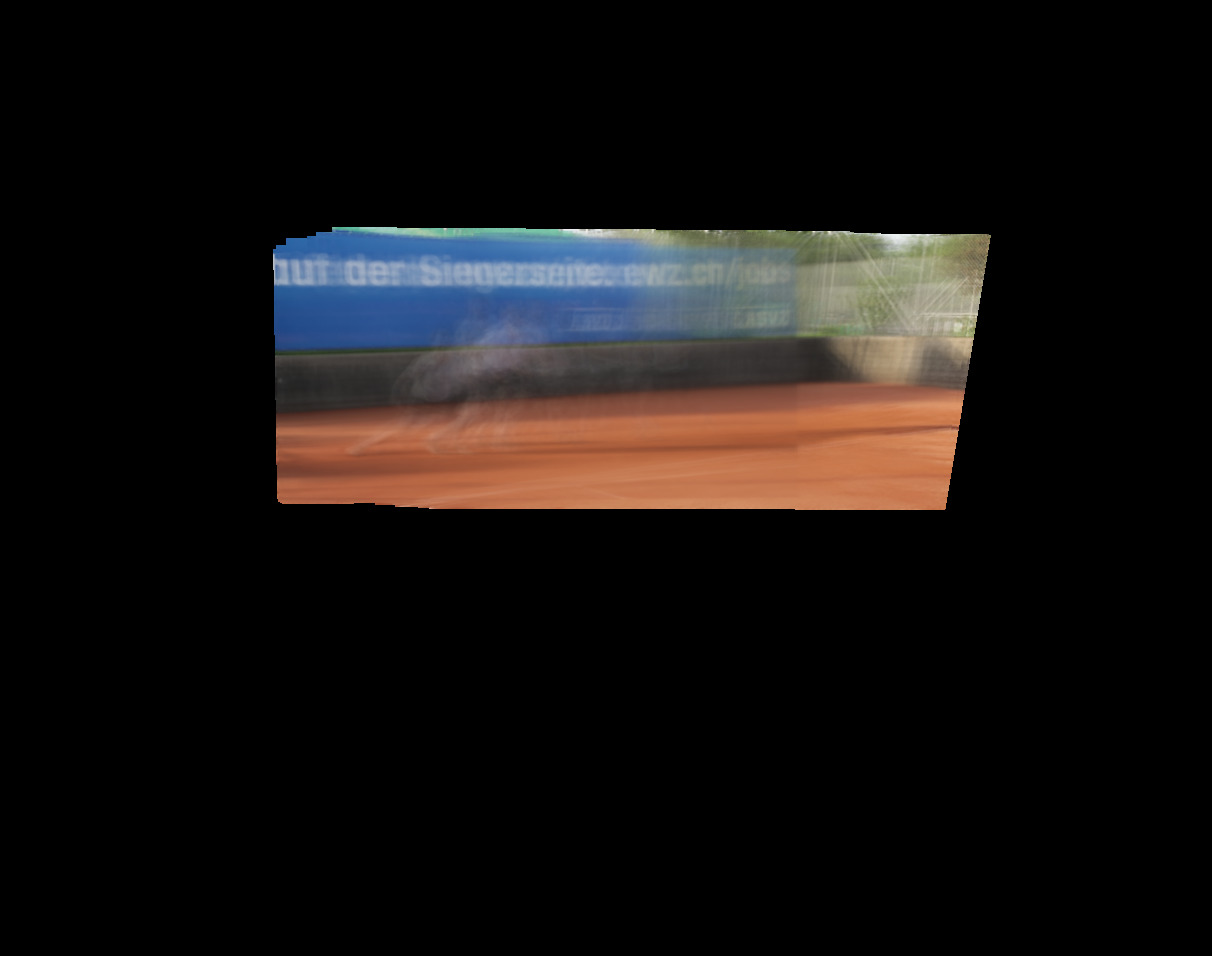}
    \includegraphics[width=\Size,trim={0mm 0mm 0mm 0mm},clip]{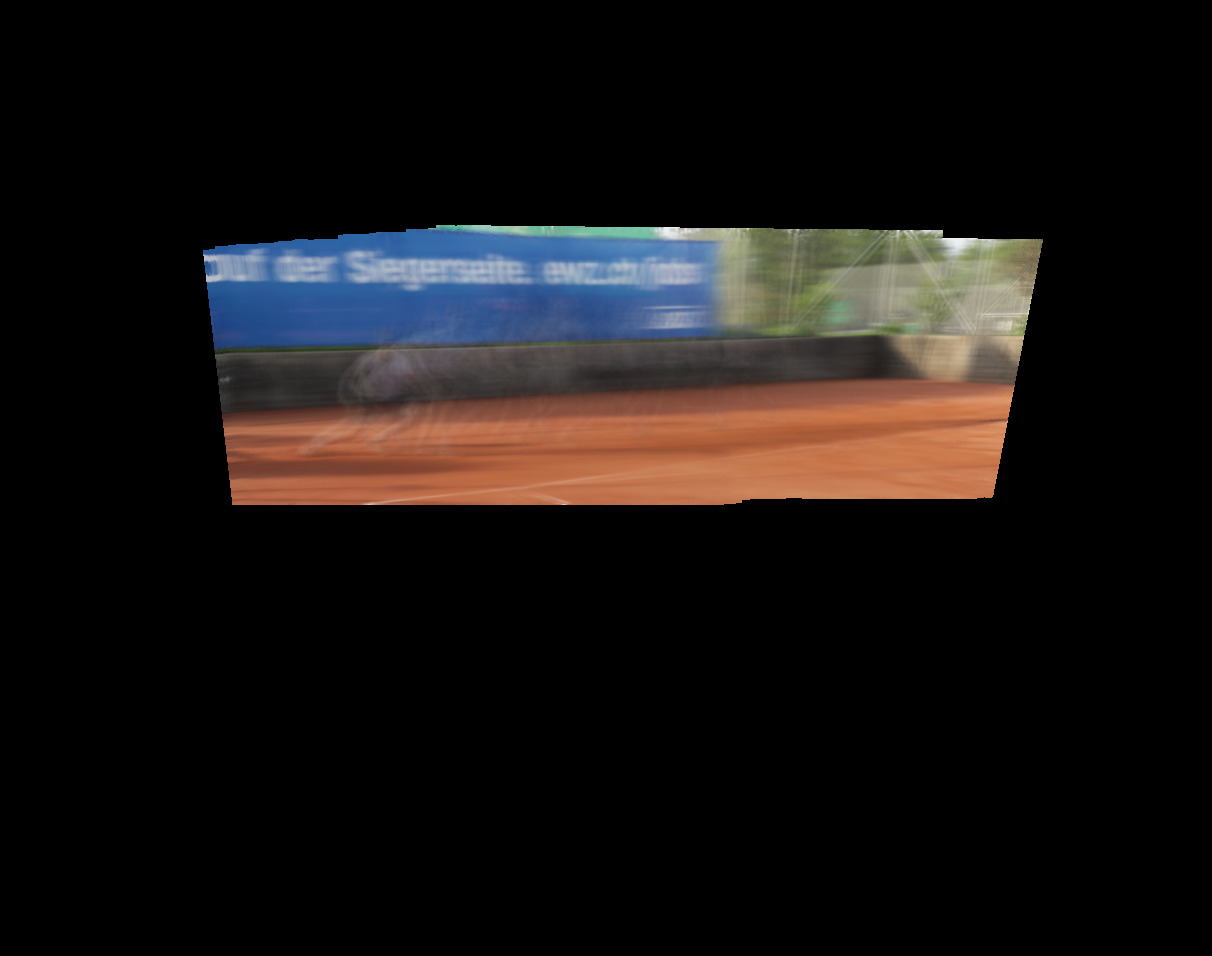}
    \includegraphics[width=\Size,trim={0mm 0mm 0mm 0mm},clip]{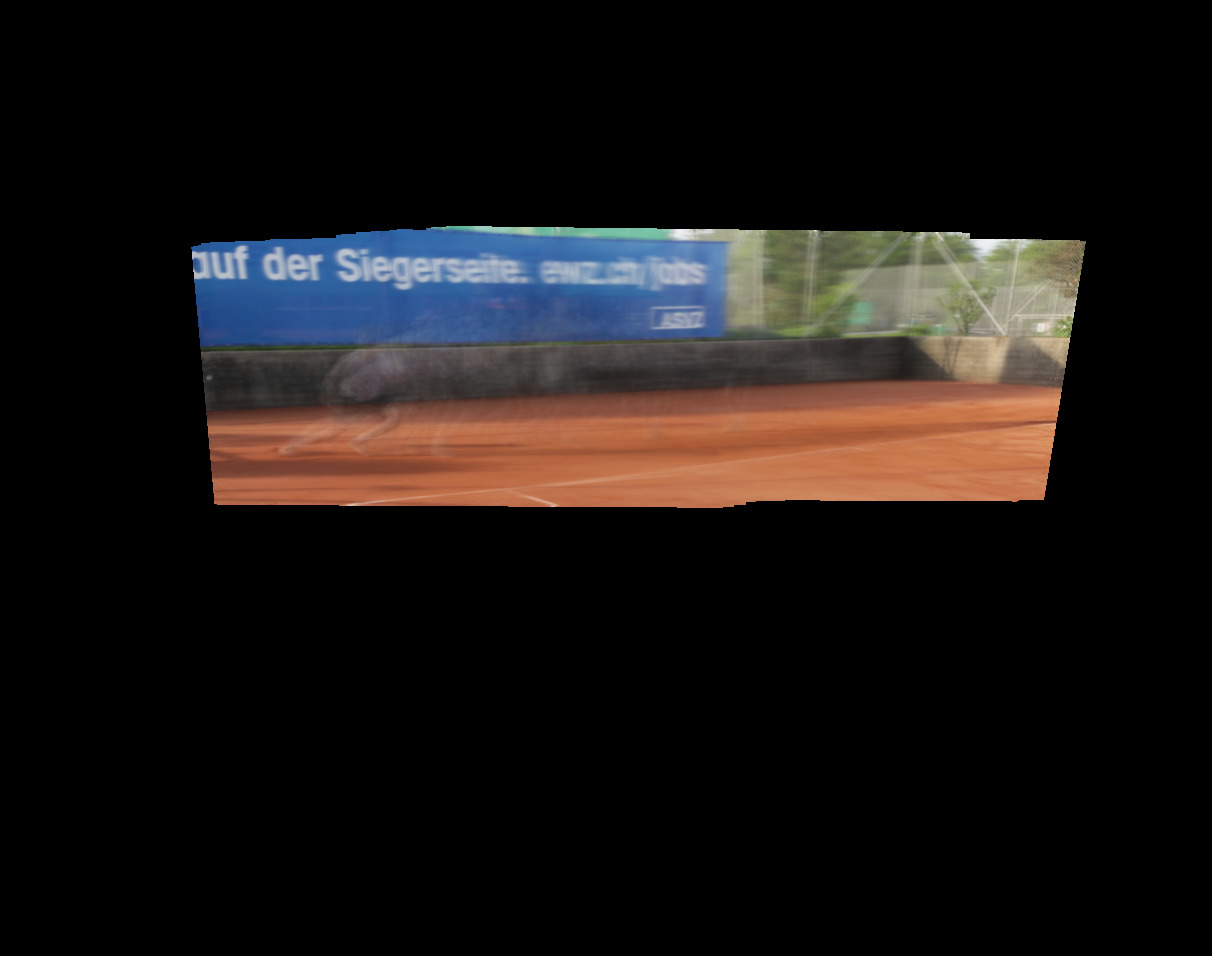}
    \includegraphics[width=\Size,trim={0mm 0mm 0mm 0mm},clip]{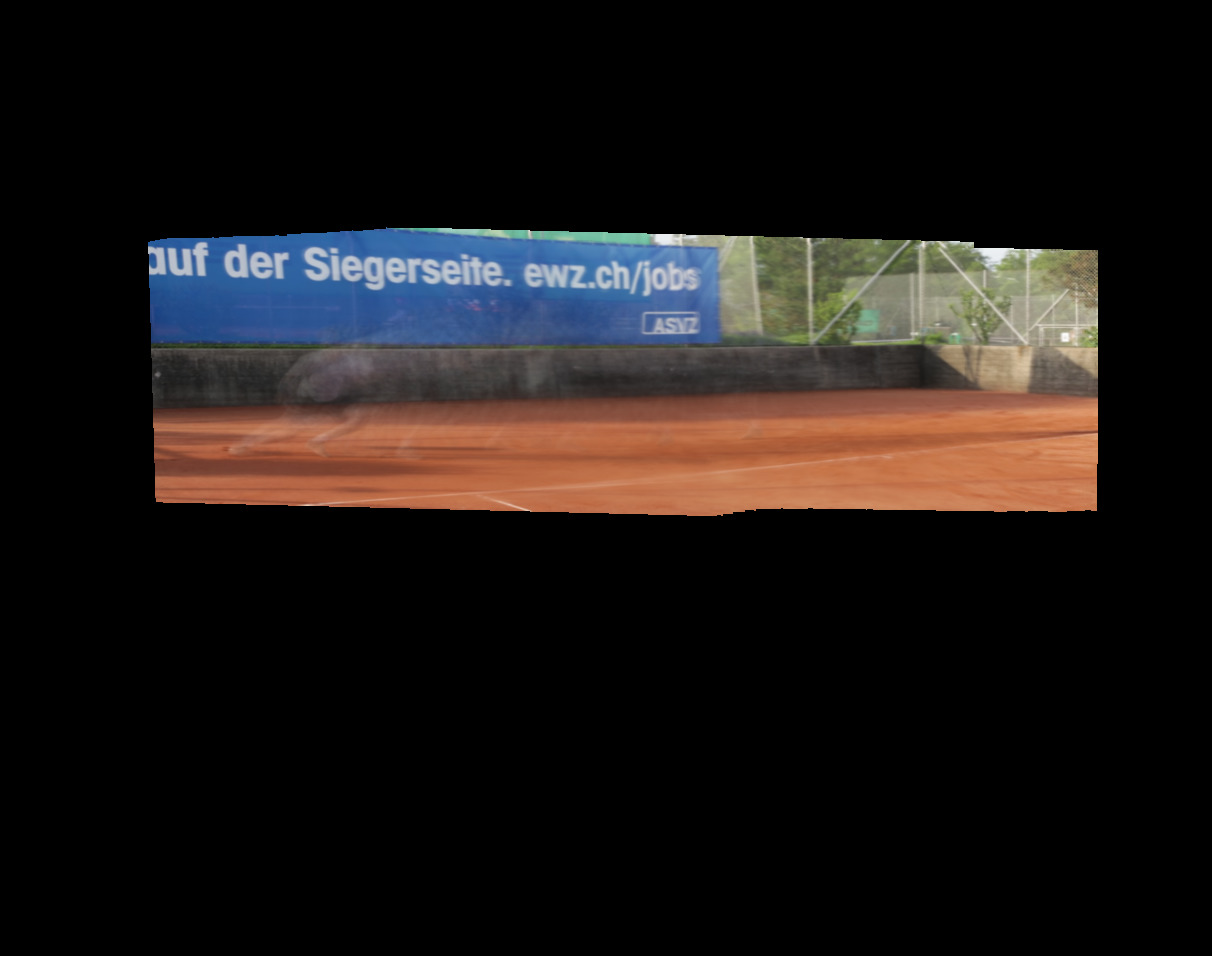}
}
\caption{Results analogous to those in~\autoref{Fig:JAproblems} except they were obtained with the proposed memory-based approach. Rightmost images are post-convergence results.}
\label{Fig:Memory}
\end{figure}

As shown in~\autoref{Fig:Memory},  the resulting targets (\ie, the $\mu$ sequence formed during the optimization) change smoothly
between epochs. 
Importantly, this behaviour has a profound and fourfold positive effect:
\textbf{1. No complicated initialization is needed}. As the optimization becomes much easier,  the initial transformations are simply taken to be the identity.
 \textbf{2. Regularization-free JA}. No form of regularization on  $(\btheta_n)_{n=1}^N$ is needed; \eg, there is no need to worry about the poor global minima from~\autoref{Sec:JAisHard}. 
  Since the optimization is gradient-based
and since each epoch lingers in the ``history'' of the process for many epochs 
before its effective weight decays to zero (due to the repeating multiplications by $\lambda\in(0,1)$), such undesired cases are eliminated altogether. 
For instance, as the stack of the original frames overlaid over each other (from the first epoch) contributes to the computation of $\mu$, either shrinking the frames to a point or moving them outside $\Omega_\mathrm{scene}$ will incur a loss.  
Our regularization-free JA is in sharp contrast to many algorithms including classical works (\eg,~\cite{Learned:PAMI:2006:DataDrivanViaJA}) 
and more recent ones (\eg,~\cite{Chelly:CVPR:2020:JA-POLS}). 
\textbf{3. Higher expressiveness.} The formulation lets us increase the expressiveness of the transformation family as needed.
For example, JA-POLS is so crucially dependent on its SE-Sync initialization and SE-based regularization, that the affine transformations it predicts
are nearly in \SEtwo\ themselves. 
In contrast, our method can not only predict 
more general transformations in the Affine group but also use broader transformation families. In our experiments we demonstrate this using the group of homographies but one may also try richer STNs such as those based on
  diffemorphisms~\cite{Skafte:CVPR:2018:DDTN,Balakrishnan:CVPR:2018:unsupervised,Dalca:NIPS:2019:learning}. 
\textbf{4. Our JA module can be used in end-to-end pipelines.}
     This is unlike not only non-DL methods but also JA-POLS~\cite{Chelly:CVPR:2020:JA-POLS} whose non-differentiable initialization prevents its JA module from being used in an end-to-end manner.   
The technical details of the training process appear in our \textbf{Supplemental Material (Supmat)}.

\subsection{Background Modeling in $\Omega$ (not $\Omega_\mathrm{scene}$)
via a Conditional Autoencoder}\label{Sec:Method:CAE} 
Upon the training of the STN, the frames become jointly aligned. 
In principle, at this point all that is left to do is to learn a background model using either non-DL methods
(\eg, based on either pixelwise mixture models or RPCA methods; see~\autoref{Sec:RelatedWork}) or deep ones (such as using a robust loss when training an autoencoder for reconstruction). 
However, there are several problems with this approach. 
First, it does not scale well: if the accumulating motion of the camera throughout the video is large, the panoramic image (of the entire scene covered throughout the video) can be huge. 
Moreover, in such a case even scalable RPCA methods will have to face an additional problem:
since the domain of each warped image captures only a small region inside the domain of the panoramic image, it means that \emph{most} of the pixels will represent missing data. 
Thus, one would need an RPCA method which can not only scale well but also succeed in situations
where more than, say,  90\%-95\% of the data is missing. 
Also important is the following. Recall that given an input image, our goal is to estimate a background image, of the same size, that corresponds to that image. Thus, why should we even bother with trying to learn a panoramic-size background model? 
In~\cite{Chelly:CVPR:2020:JA-POLS}, the discussion above motivated the learning of multiple local RPCA models and then, for estimating the background of a given image, only a subset of those models whose domains overlapped with the frame of interest were used. That solution, however, means that the number of models to be learned grows with the size of the panoramic image. Moreover, its non-DL formulation was another reason why JA-POLS was not an end-to-end method.

\begin{figure}[!t]
    \centering
    \includegraphics[width=1\linewidth,trim={0mm 0mm 0mm 18mm},clip]{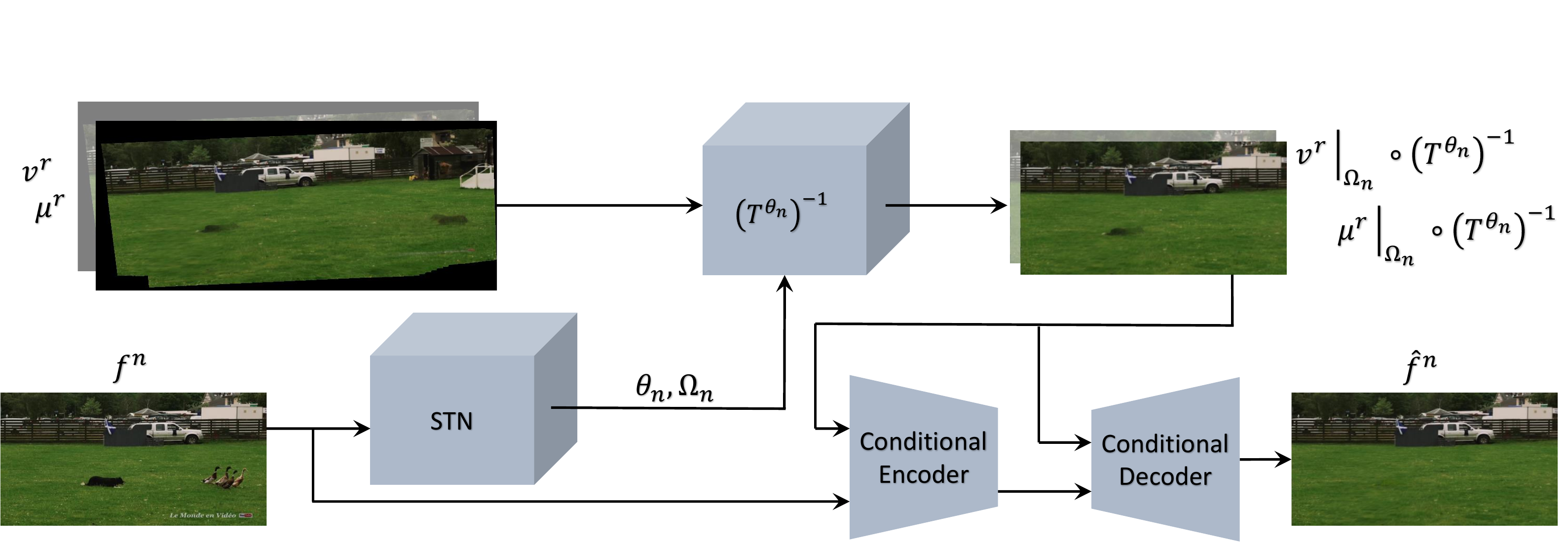}
    \captionsetup{justification=raggedright, singlelinecheck=false}
    \caption{The background-modeling module.
    After the STN module was trained using~\autoref{Alg:Update_target_and_compute_batch_loss},
    the robust panoramic moments, $\mu^r$ and $v^r$, are computed.
A CAE is trained for a robust reconstruction task, using the  transformation parameters, $(\btheta_n)_{n=1}^N$, estimated by the (frozen) STN. The CAE's output is $\widehat{f^n}$, the estimated background
associated with $f^n$ and the conditioning  is done by (un)warping $\mu^r$ and $v^r$  towards each input training image, $f^n$. 
 During test time the process is similar,
except that the transformation being used is the one predicted
by the STN.     }
    \label{Fig:Pipeline}
    \end{figure}

Here we propose a better alternative, whose pipeline is summarized in \autoref{Fig:Pipeline}: use a CAE to learn a  background model 
whose domain is small. This has two advantages:
1) It does not compromise the end-to-end nature of the method. 
2) We need to learn only a single model (unlike in~\cite{Chelly:CVPR:2020:JA-POLS}) and its domain is small, fixed, and does not grow with the size of panoramic image (unlike in PRPCA). 
Concretely, rather than learning a background model (or models) whose domain is $\Omega_\mathrm{scene}$, we train a CAE on the original (\ie, non-warped) input frames, using
a robust reconstruction error and, for each input frame $f^n$, conditioning  both the encoder and the decoder on (robust versions of) the mean and variance of the pixel stacks, but not before unwarping those central moments from $\Omega_\mathrm{scene}$ towards $f^n$.  
We now provide the details. 
The first and second central moments, denoted by
$\mu^{r}$
and 
$v^{r}$,
respectively, 
are $C$-channel $H\times W$ images defined on $\Omega_\mathrm{scence}$
and computed rubustly using trimmed averaging as follows. 
Fix $\bx\in\Omega_\mathrm{scene}$, let $N_\bx=|\set{n:M^n_\bx>0}|$,
and let $(g^{(1)}_{\bx,c},\ldots, g^{(N_{\bx})}_{\bx,c})$ be the order statistics of 
$p_{\bx,c}$. 
The values of 
$\mu^{r}$ and $v^{r}$
at $\bx$ in channel $c$ are computed, respectively, by 
\begin{align}
\mu^{r}_{\bx,c} = \tfrac{1}{(1-2 \alpha)  N_\bx}\sum_{i={ \alpha N_\bx}}^{(1-\alpha)N_\bx}
g^{(i)}_{\bx,c}
& & 
v^{r}_{\bx,c} = \tfrac{1}{(1-2 \alpha)  N_\bx}\sum_{i={ \alpha N_\bx}}^{(1-\alpha)N_\bx}
(g^{(i)}_{\bx,c}-\mu^{r}_{\bx,c})^2 \, . 
\end{align}
Such trimmed averaging is a standard technique for computing robust moments~\cite{Hauberg:CVPR:2014:TGA}. The trimming parameter, $\alpha$,  was empirically set to $\alpha=0.3$ as it provided a good balance between sample size and robustness.  That said, the results when using any other value in the wide range
between 20\% and almost 50\% were similar. 
Next, when $f^n$ is fed into the CAE,  the encoder and the decoder 
are conditioned by 
\begin{align}
\mu^n\triangleq 
 \left(\left. \mu^{r}\right|_{\Omega_n}\right)
\circ (T^{\btheta_n})^{-1}
 \text{ and } 
 v^n\triangleq
 \left(\left. v^{r}\right|_{\Omega_n}\right)
 \circ (T^{\btheta_n})^{-1}
\end{align}
which are $h\times w$ images (with $C$ channels) defined on $\Omega$ and are nothing more than the portion of 
$\mu^{r}$ and $v^{r}$
that is relevant for $f^n$. 
Using a code whose length was only 4, 
the CAE  was trained with the following loss: 
\begin{align}
 \Lcal_{AE} &=  \sum\nolimits_{n=1}^N \sum\nolimits_{c=1}^C\sum\nolimits_{\bx'\in\Omega}\rho_\mathrm{recon}(f^n_{\bx',c}-\widehat{f}^n_{\bx',c}) \\
 \widehat{f}^n &=  \mathrm{Decoder}(\mathrm{Encoder}(f^n;
 \mu^n
 ,v^n
);\mu^n,v^n)
\end{align}
where 
$\widehat{f}^n $ is the output of the CAE and 
$\rho_\mathrm{recon}$ is a differentiable robust error function. 
We remark that, by design, the fact that $\mu^n$ and $v^n$ are of the same dimensions
as the input, $f^n$, also means it is easy to implement the conditioning via a convolutional layer. 
For more details about the CAE (whose architecture is based on the AE from~\cite{Balle:ICLR:2017:AE}) as well as other training details, 
see our \textbf{Supmat}. 
Finally, $\rho_\mathrm{recon}$
should usually be more robust than $\rho_{\mathrm{JA}}$. 
The reason is that  while in JA the influence of foreground objects is relatively small,
in the CAE-based reconstruction
it is important, in every pixel, to eliminate the outliers
(\ie, the foreground pixels) as much as possible. 
Thus, we use the  smoothed $\ellOne$ loss (which is closely-related
to Huber's function~\cite{Black:IJCV:1996:Robust})
for $\rho_\mathrm{JA}$
and the Geman-McClure error function~\cite{GemanMcClure:BISI:1987} for $\rho_\mathrm{recon}$. 
See \textbf{Supmat} for details. 

  \section{Results}\label{Sec:Results}
We experimented with 4 variants of  the proposed DeepMCBM:
1. \textbf{Basic/Aff}: This version uses only the STN-based JA module, without the CAE.
It estimates the background by simply unwarping the
robust mean towards the input image. The transformations
used in the STN belong to the Affine group (the invertibility of the transformations was guaranteed via the matrix exponential; see \textbf{Supmat}). 
2. \textbf{CAE/Aff}: This version too uses the Affine STN
 but also uses the CAE module (for estimating the background). 
3. \textbf{Basic/Hom}
and 4. \textbf{CAE/Hom}: Similar to Basic/Aff and  CAE/Aff, respectively, except that homographies are used instead of affine transformations. 
We compared those 4 variants  with several methods:
PRPCA~\cite{Moore:TCI:2019:PRPCA};
JA-POLS~\cite{Chelly:CVPR:2020:JA-POLS};
PanGAEA~\cite{Gilman:ICCVw:2019:Panoramic};
DECOLOR~\cite{zhou:2012:DECOLOR};
PCP\_PTI~\cite{Chau:2017:incPCP_PTI};
PRAC~\cite{Guo:2013:PRAC}. 
The 13 videos that we tested on are ones typically used for evaluation of methods in this area and are taken from well-known datasets~\cite{Wang:CVPRw:2014:cdnet,Pont-Tuset:2017:benchmark_davis}. 
Those movies cover camera motions in a variety of types, sizes,
speed, zoom changes, \etc.  
It should be noted that, due to their scalability limitaitons,  PRPCA
and PanGAEA could not run on
the \textbf{ContinuousPan} video as the covered scene in the latter
was too large. JA-POLS failed running on 
\textbf{zoomInZoomOut} (the significant zoom changes broke its key assumption). 
\autoref{Fig:ResultsVisualComparison} contains a visual comparison, on select example videos, of DeepMCBM (in its CAE/Hom variant), PRPCA, JA-POLS, and PanGAEA. 
Results of the other (and less successful) methods (DECOLOR; PCP\_PTI; PRAC),
as well as more visual results (including videos) 
are in the \textbf{Supmat}.

Given  an estimate of the background, subtracting it from the original frame yields a difference that can serve to determine foreground/background separation. 
To quantify the results in a threshold-independent way, for each method and each video we computed the Receiver Operating Characteristic (ROC) curve (using the ground truth) and its Area Under the Curve (AUC). 
The  ROC curves 
are included in \textbf{Supmat}.  
We emphasize that our method is unsupervised and the ground truth information was used only for evaluation. 
\autoref{Table:AUC}, summarizing  the AUC results,
shows that DeepMCBM, especially with its
CAE variants, is, overall, the leading method. In cases where DeepMCBM is not the first it is typically the runner-up. 
Moreover, unlike some competitors, DeepMCBM was applicable in all cases considered.
The visual examples also illustrate how the CAE helps achieving a better
estimate of the background than that one obtained by merely using the unwarped $\mu^r$. 
We remark that our fixed code size, 4,  is so small since: 1) the
goal is not a typical reconstruction but to filter out foreground objects; 2) our AE is conditional so it is unsurprising
a small size suffices. We could have made the code size
video-dependent and thus improve results even further, but
felt that a fixed size is simpler and makes a comparison 
with other methods fairer.

\textbf{Predicting background for previously-unseen misaligned frames. }
In the comparison above, we focused on background/foreground estimation
in the input videos on which the competing models (ours included) were learned. 
However, like JA-POLS, but unlike all the other methods, our method can predict the background in  frames that were not included in the learning
(more accurately, some of the  competing methods can predict the background
in the next constitutive frame, but they are
 unable to do so for misaligned frames in general such as those that are not consecutive).
Due to space limits, we demonstrate that capability of DeepMCBM in the~\textbf{Supmat}. 

\newcommand{\W}{.2}
\newcommand{\dummyfig}[1]{
  \noindent
  \raisebox{-0.0ex}
{
    \begin{minipage}[c][3ex][c]{3em}
      \centering{#1}
    \end{minipage}
  }
}
\begin{figure}[H]
    \begin{tabular}{|c|c|c|c|c|c|}
        \hline
        \includegraphics[width=.2\linewidth,valign=m]{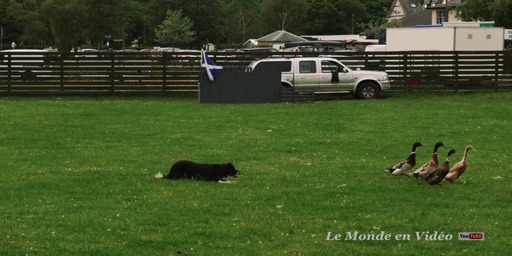} & 
        \includegraphics[width=.2\linewidth,valign=m]{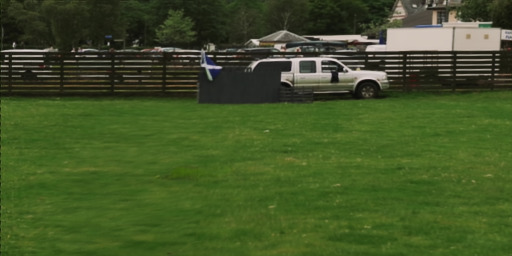} & 
        \includegraphics[width=.2\linewidth,valign=m]{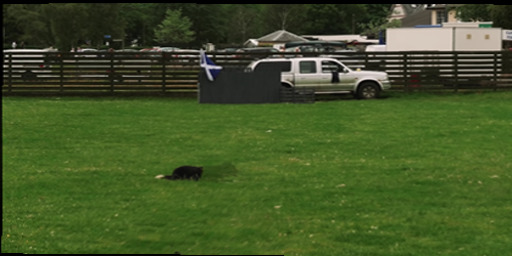} & 
        \includegraphics[width=.2\linewidth,valign=m]{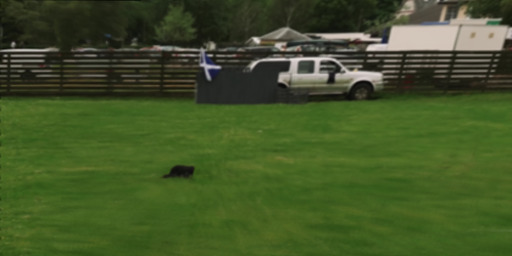} & 
        \includegraphics[width=.2\linewidth,valign=m]{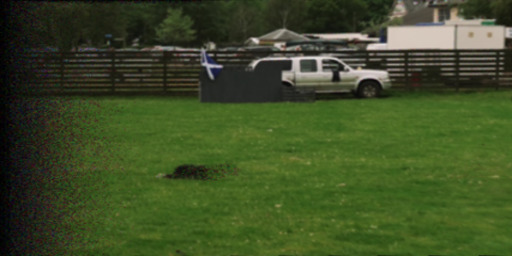}\\
        \includegraphics[width=.2\linewidth,valign=m]{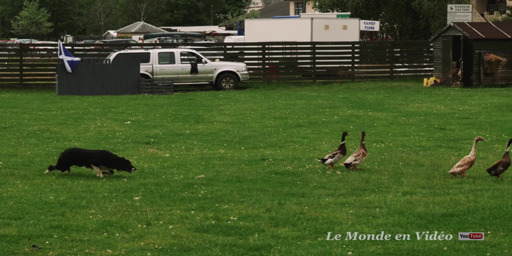} & 
        \includegraphics[width=.2\linewidth,valign=m]{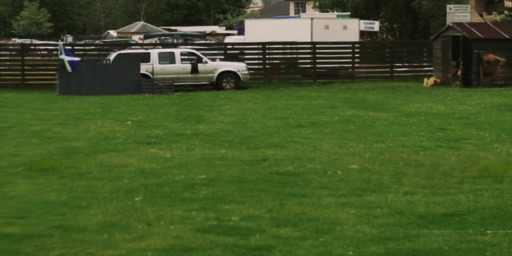} & 
        \includegraphics[width=.2\linewidth,valign=m]{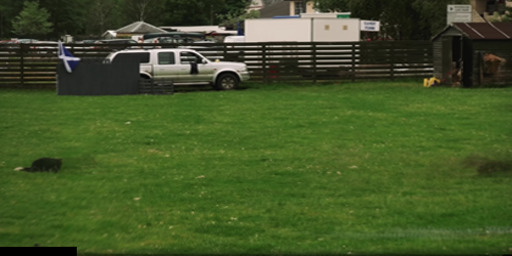} & 
        \includegraphics[width=.2\linewidth,valign=m]{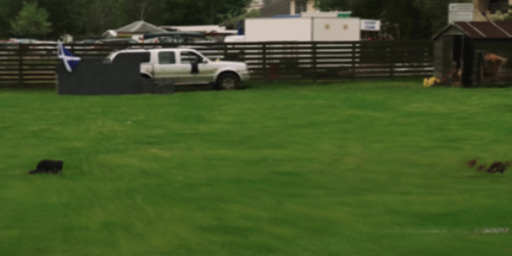} & 
        \includegraphics[width=.2\linewidth,valign=m]{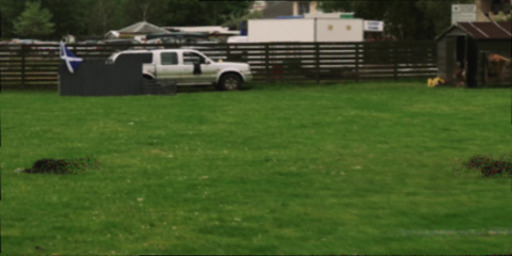}\\
        \includegraphics[width=.2\linewidth,valign=m]{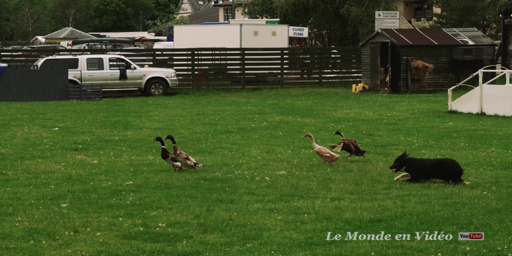} & 
        \includegraphics[width=.2\linewidth,valign=m]{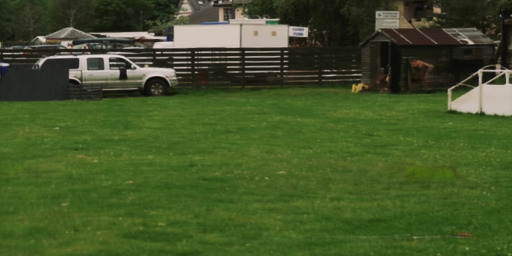} & 
        \includegraphics[width=.2\linewidth,valign=m]{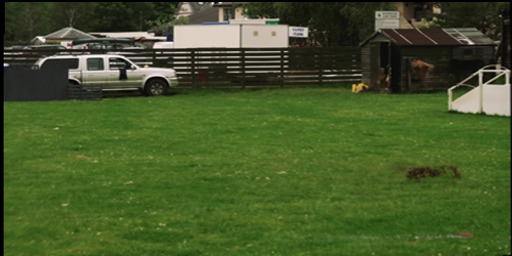} & 
        \includegraphics[width=.2\linewidth,valign=m]{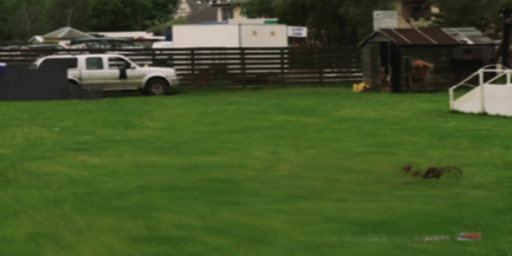} & 
        \includegraphics[width=.2\linewidth,valign=m]{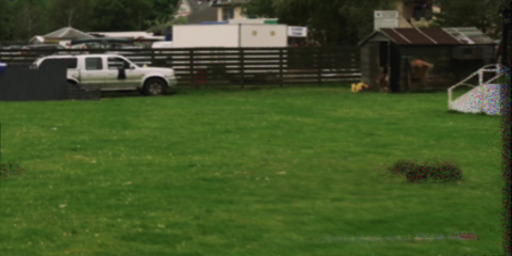}\\
        \hline
        \hline
        \includegraphics[width=.2\linewidth,valign=m]{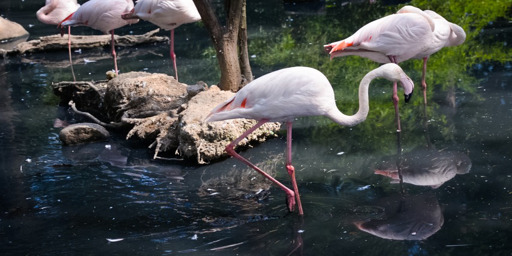} & 
        \includegraphics[width=.2\linewidth,valign=m]{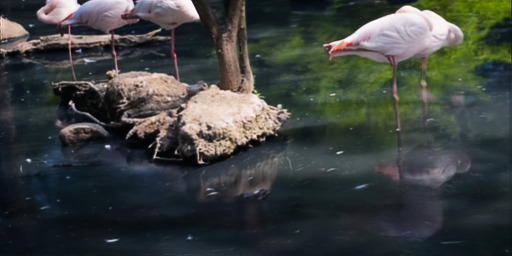} & 
        \includegraphics[width=.2\linewidth,valign=m]{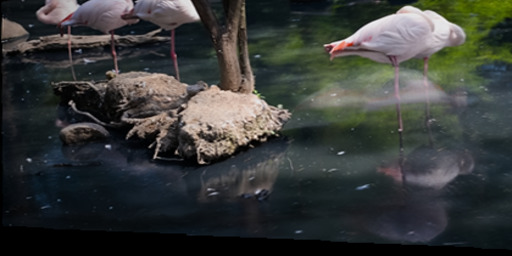} & 
        \includegraphics[width=.2\linewidth,valign=m]{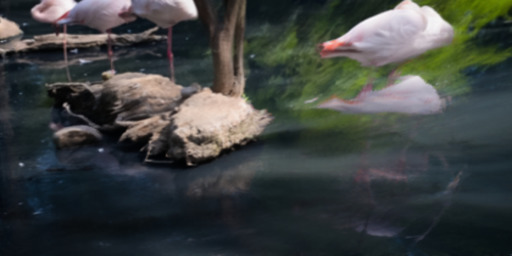} & 
        \includegraphics[width=.2\linewidth,valign=m]{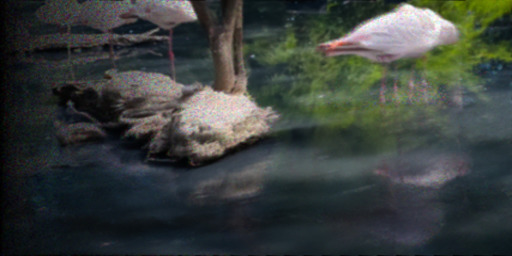}\\
        \includegraphics[width=.2\linewidth,valign=m]{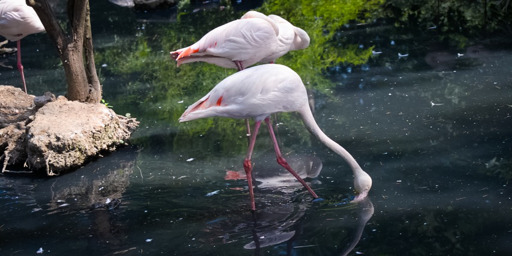} & 
        \includegraphics[width=.2\linewidth,valign=m]{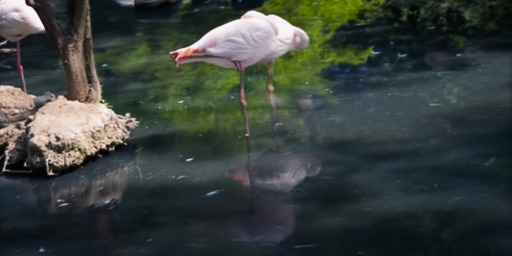} & 
        \includegraphics[width=.2\linewidth,valign=m]{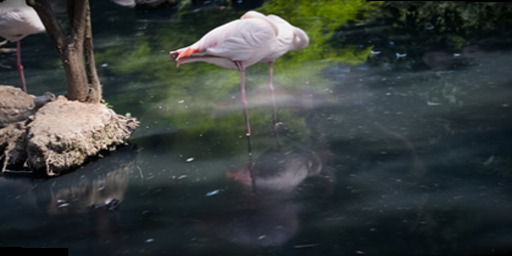} & 
        \includegraphics[width=.2\linewidth,valign=m]{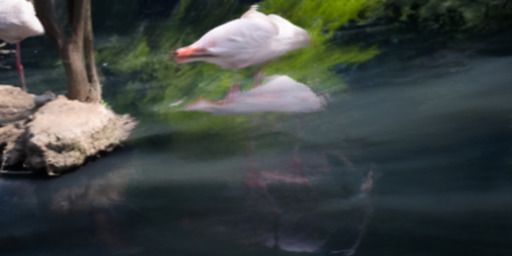} & 
        \includegraphics[width=.2\linewidth,valign=m]{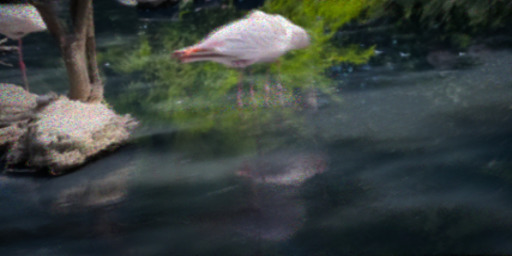}\\
        \includegraphics[width=.2\linewidth,valign=m]{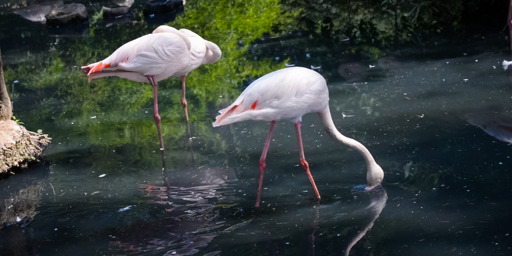} & 
        \includegraphics[width=.2\linewidth,valign=m]{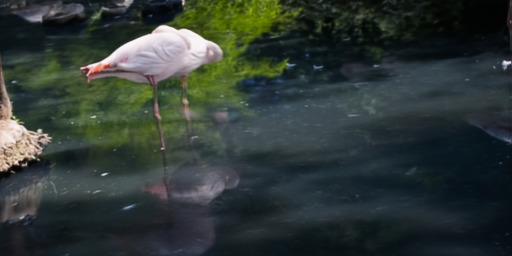} & 
        \includegraphics[width=.2\linewidth,valign=m]{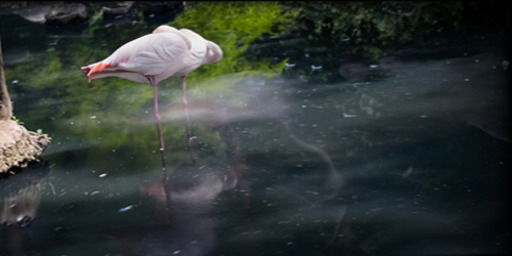} & 
        \includegraphics[width=.2\linewidth,valign=m]{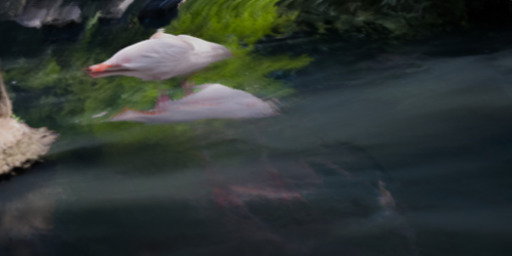} & 
        \includegraphics[width=.2\linewidth,valign=m]{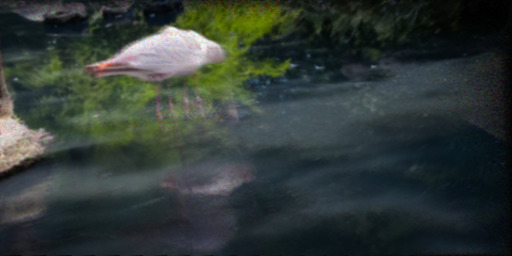}\\
        \hline
        \hline
        \includegraphics[width=.2\linewidth,valign=m]{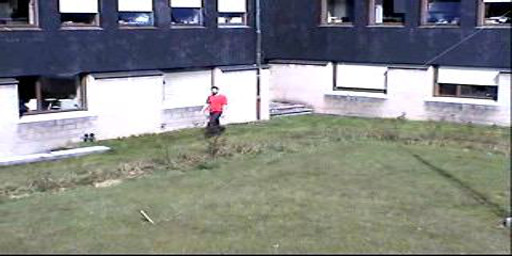} & 
        \includegraphics[width=.2\linewidth,valign=m]{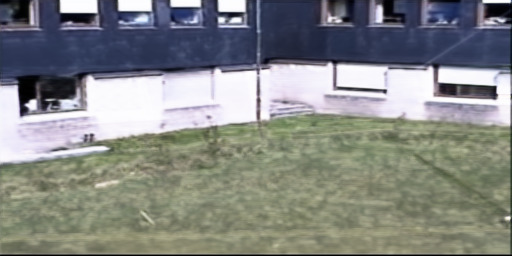} & 
        \dummyfig{}&
        \includegraphics[width=.2\linewidth,valign=m]{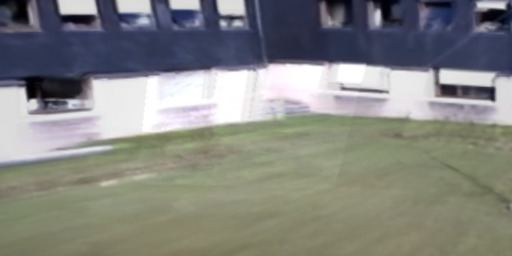} & 
        \includegraphics[width=.2\linewidth,valign=m]{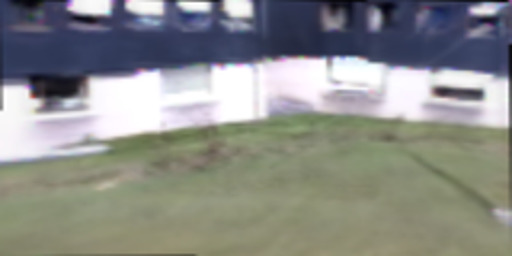}\\
        \includegraphics[width=.2\linewidth,valign=m]{./fig_compare_results_imgs/zoomInZoomOut/input/in000401_resize.jpg} & 
        \includegraphics[width=.2\linewidth,valign=m]{./fig_compare_results_imgs/zoomInZoomOut/Ours/frame000400_resize.jpg} & 
        \dummyfig{N/A (1)}&
        \includegraphics[width=.2\linewidth,valign=m]{./fig_compare_results_imgs/zoomInZoomOut/PRPCA/f_400_resize.jpg} & 
        \includegraphics[width=.2\linewidth,valign=m]{./fig_compare_results_imgs/zoomInZoomOut/PanGAEA/f_0400_resize.jpg}\\
        \includegraphics[width=.2\linewidth,valign=m]{./fig_compare_results_imgs/zoomInZoomOut/input/in000401_resize.jpg} & 
        \includegraphics[width=.2\linewidth,valign=m]{./fig_compare_results_imgs/zoomInZoomOut/Ours/frame000400_resize.jpg} & 
        \dummyfig{}&
        \includegraphics[width=.2\linewidth,valign=m]{./fig_compare_results_imgs/zoomInZoomOut/PRPCA/f_400_resize.jpg} & 
        \includegraphics[width=.2\linewidth,valign=m]{./fig_compare_results_imgs/zoomInZoomOut/PanGAEA/f_0400_resize.jpg}\\
        \hline
        \hline
        \includegraphics[width=.2\linewidth,valign=m]{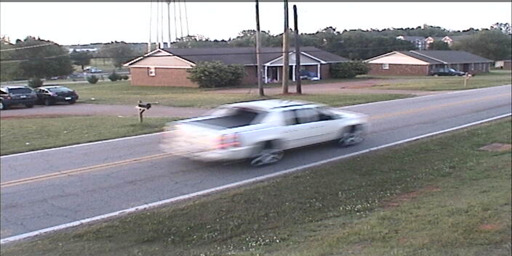} & 
        \includegraphics[width=.2\linewidth,valign=m]{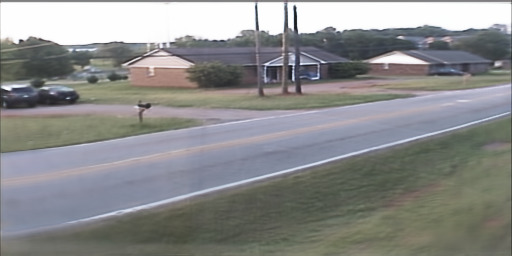} & 
        \includegraphics[width=.2\linewidth,valign=m]{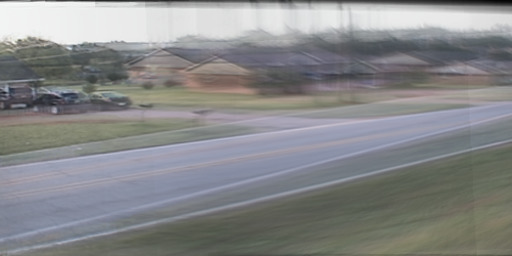} & 
        \dummyfig{} & 
        \dummyfig{}\\        
        \includegraphics[width=.2\linewidth,valign=m]{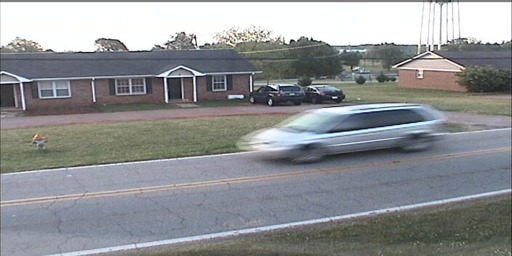} & 
        \includegraphics[width=.2\linewidth,valign=m]{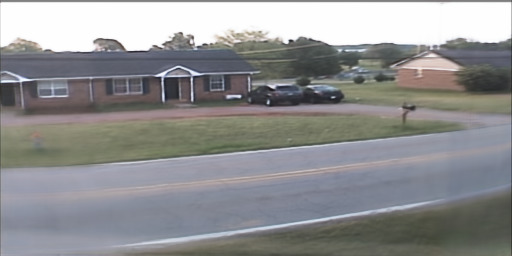} & 
        \includegraphics[width=.2\linewidth,valign=m]{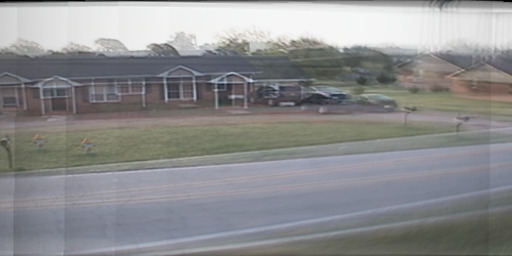} & 
        \dummyfig{N/A (2)} & 
        \dummyfig{N/A (3)}\\        
        \includegraphics[width=.2\linewidth,valign=m]{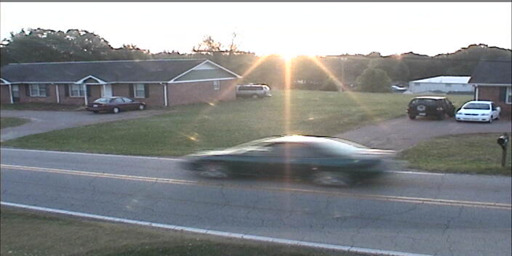} & 
        \includegraphics[width=.2\linewidth,valign=m]{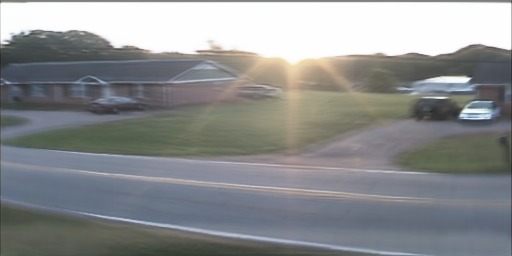} & 
        \includegraphics[width=.2\linewidth,valign=m]{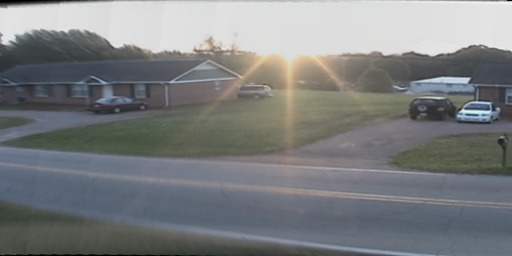} & 
        \dummyfig{} & 
        \dummyfig{}\\
        \hline
        \hline                
        Input & Ours & JA-POLS & PRPCA & PanGAEA \\
        \hline       
    \end{tabular}
\caption{Visual Comparison: Select Results. Note the ghosting artifacts. N/A (1) large zoom changes failed JA-POLS completely, N/A (2) out-of-memory on a 256GB RAM machine, N/A (3) failed to run: Matlab process was killed  }
\label{Fig:ResultsVisualComparison_2}

\end{figure}

\textbf{Ablation Study.}
As~\autoref{Table:AUC} shows, the AE usually improves performance. In particular, its role is especially important when a foreground object spends a long portion of time in a static position  (\eg, the dog in the dog-gooses or the flamingo). In such cases, the robust mean alone still tends to capture some ``ghosting'' artifacts (as usually do all the competing methods) while the CAE helps correctly identifying that object as belonging to the foreground. 
The importance of the memory-based approach was also demonstrated in~\autoref{Fig:JAproblems} and~\autoref{Fig:Memory}. In particular, the JA failures
in~\autoref{Fig:JAproblems} imply that no subsequent background model could be built there, making a quantitative comparison (between using the memory term and not using it) a moot point.  
Finally, note that a basic (\ie, unconditional AE) that knows nothing 
about the alignment has no chance here as it can only either simply reconstruct the entire frames (\ie, with the undesired foreground objects) or fail in the reconstruction. Thus, when simply dropping the conditioning from our CAE, the resulting AE fails badly in  background modeling; 
\eg, its AUC for the Tennis video is 0.701 while DeepMCBM's AUC score is 0.963.

  \begin{table}[t]
  \resizebox{\textwidth}{!}{%
  \begin{tabular}{lcccccccccc}
  \hline
  \multicolumn{1}{|l|}{} &
    \multicolumn{4}{c|}{DeepMCBM (Ours)} &
    \multicolumn{1}{c}{} &
    \multicolumn{1}{c}{} &
    \multicolumn{1}{c}{} &
    \multicolumn{1}{c}{} &
    \multicolumn{1}{c}{} &
    \multicolumn{1}{c|}{} \\ \hline
  \multicolumn{1}{|l|}{Sequence} &
    \multicolumn{1}{c|}{Basic/Aff} &
    \multicolumn{1}{c|}{CAE/Aff} &
    \multicolumn{1}{c|}{Basic/Hom} &
    \multicolumn{1}{c|}{CAE/Hom} &
    \multicolumn{1}{c|}{~\cite{Moore:TCI:2019:PRPCA}} &
    \multicolumn{1}{c|}{~\cite{Gilman:ICCVw:2019:Panoramic}} &
    \multicolumn{1}{c|}{~\cite{Chau:2017:incPCP_PTI}} &
    \multicolumn{1}{c|}{~\cite{Chelly:CVPR:2020:JA-POLS}} &
    \multicolumn{1}{c|}{~\cite{Guo:2013:PRAC}} &
    \multicolumn{1}{c|}{~\cite{zhou:2012:DECOLOR}} \\ \hline
  \textbf{bmx-trees}        & .898 & .896          & .916 & .908          & .894          & .786          & .837 & \textbf{.930} & .664 & .737          \\
  \textbf{boxing-fisheye}   & .924 & .893          & .927 & .898          & \textbf{.935} & .932          & .728 & .892          & .627 & .763          \\
  \textbf{breakdance-flare} & .931 & .933          & .953 & .963          & .960          & \textbf{.972} & .740 & .897          & .806 & .667          \\
  \textbf{continuousPan}    & .897 & \textbf{.940} & .895 & .938          & N/A           &  N/A          & .846 & .449           & .656 & .760          \\
  \textbf{dog-gooses}       & .954 & \textbf{.984} & .955 & \textbf{.984} & .942          & .917          & .721 & .947          & .747 & .886          \\
  \textbf{flamingo}         & .962 & \textbf{.980} & .961 & \textbf{.980} & .891          & .957          & .638 & .947          & .560 & .656          \\
  \textbf{horsejump-high}   & .932 & .942          & .932 & .943          & \textbf{.958} & .908          & .783 & .914          & .713 & .892          \\
  \textbf{sidewalk}         & .886 & .908          & .889 & .932          & .812          & .702          & .635 & .851          & .780 & \textbf{.935} \\
  \textbf{stroller}         & .877 & .885          & .740 & .756          & .762          & \textbf{.904} & .594 & .807          & .613 & .721          \\
  \textbf{stunt}            & .963 & \textbf{.979} & .961 & .978          & .959          & .954          & .899 & .930          & .711 & .781          \\
  \textbf{swing}            & .880 & .877          & .887 & .897          & \textbf{.942} & .879	         & .805 & .874          & .722 & .812          \\
  \textbf{tennis}           & .960 & .961          & .959 & \textbf{.963} & .943          & .929          & .831 & .932          & .787 & .852          \\
  \textbf{zoomInZoomOut}    & .981 & \textbf{.994} & .981 & \textbf{.994} & .979          & .958          & .720 & N/A           & .885 & .957         
  
  \\
  \bottomrule     
  \end{tabular}%
}
\caption{AUC scores for each method on each sequence.}     \label{Table:AUC}
\end{table}

\section{Conclusion}\label{Sec:Conclusion}
The proposed DeepMCBM is an end-to-end DL solution for modeling background
in a video from a moving camera. It supports a wide range of camera-motion
types and sizes, scales gracefully, and achieves SOTA results. 
While we experimented with either affine transformations or homographies, 
DeepMCBM also supports more expressive transformations.  
The proposed regularization-free STN-based JA strategy may find usage in other applications, thereby the potential impact of this work  may be broader than MCBMs. One limitation of our work
is that, since DL involved, the training is slower in comparison to some competitors (JA-POLS excluded). However, we believe the SOTA results
together with the other benefits DeepMCBM brings
(end-to-end; scalability;
the ability to predict background for previously-unseen misaligned frames; \etc)
justifies it. The main failure case of the method is when foreground objects are large
and much closer to the camera than the background is.

{\textbf{Acknowledgements.}
This work was supported in part by the Lynn and William Frankel Center at BGU CS and by Israel Science Foundation Personal Grant \#360/21. G.E.~was also funded by the VATAT National excellence scholarship for female
Master’s students in Hi-Tech-related fields.}

\clearpage
%
%
\bibliographystyle{splncs04}
\bibliography{./refs}
\end{document}


\pagestyle{headings}
\mainmatter
\def\ECCVSubNumber{4790}  

\title{An End-to-end Moving Camera Background Model
\newline ------------------\newline
Supplemental Material}
\titlerunning{A Deep Moving-camera Background Model}
%
\author{Guy Erez\inst{1}\orcidID{0000-0002-4545-6664} \and
Ron Shapira Weber\inst{1}\orcidID{0000-0003-4579-0678} \and
Oren Freifeld\inst{1}\orcidID{0000-0001-9816-9709}}
\authorrunning{Erez et al.}
%
\institute{Ben-Gurion University of the Negev, Be'er Sheva, Israel \\
\email{\{ergu,ronsha\}@post.bgu.ac.il, orenfr@cs.bgu.ac.il}}
\maketitle
\begin{abstract}
 The supplemental material includes, in addition to this document, 
select examples of \textbf{videos showing the results of our method next to the original videos}. 
Additionally, and due to space limits, additional visual comparisons between the different methods (using PDF presentations so it be will easy to browse back and forth between the frames) are available at \url{https://github.com/BGU-CS-VIL/DeepMCBM}.  

As for \textbf{this document}, it contains the following:
\begin{enumerate}
 \item A comparison of ROC curves of different methods on various videos.
      These curves correspond to the AUC values reported in Table 1 in the paper.
 \item A demonstration of DeepMCBM's capability of predicting 
 the background in previously-unseen misaligned frames from the video it was trained on (a capability lacking in most other methods, except JA-POLS~\cite{Chelly:CVPR:2020:JA-POLS}).  
 \item The details of the robust error functions that we used. 
  \item The technical details of the evaluation procedures.      
\item The technical details of the architecture and training process.      
\item An explanation how, in the affine Spatial Transformer Net, 
the invertibility of the affine transformations
is guaranteed via the matrix exponential. 
\end{enumerate}


\end{abstract}
\section{ROC Curves (Related to the Values from Table 1 in the Paper)}
\begin{figure}[H]
\newcommand{\myW}{0.27\linewidth}
    \centering
    \includegraphics[width=\myW,trim={8mm 8mm 10mm 10mm},clip]{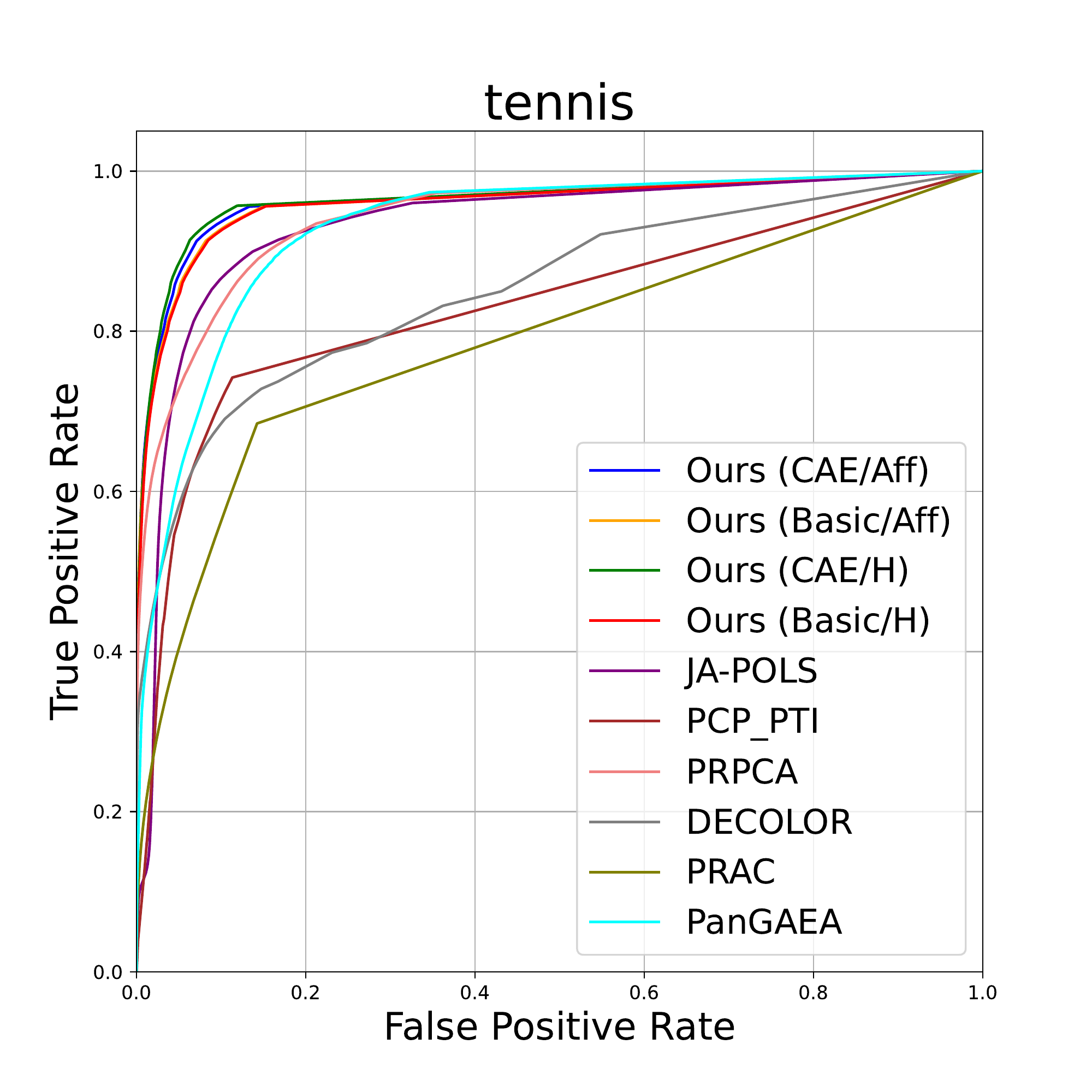}
    \includegraphics[width=\myW,trim={8mm 8mm 10mm 10mm},clip]{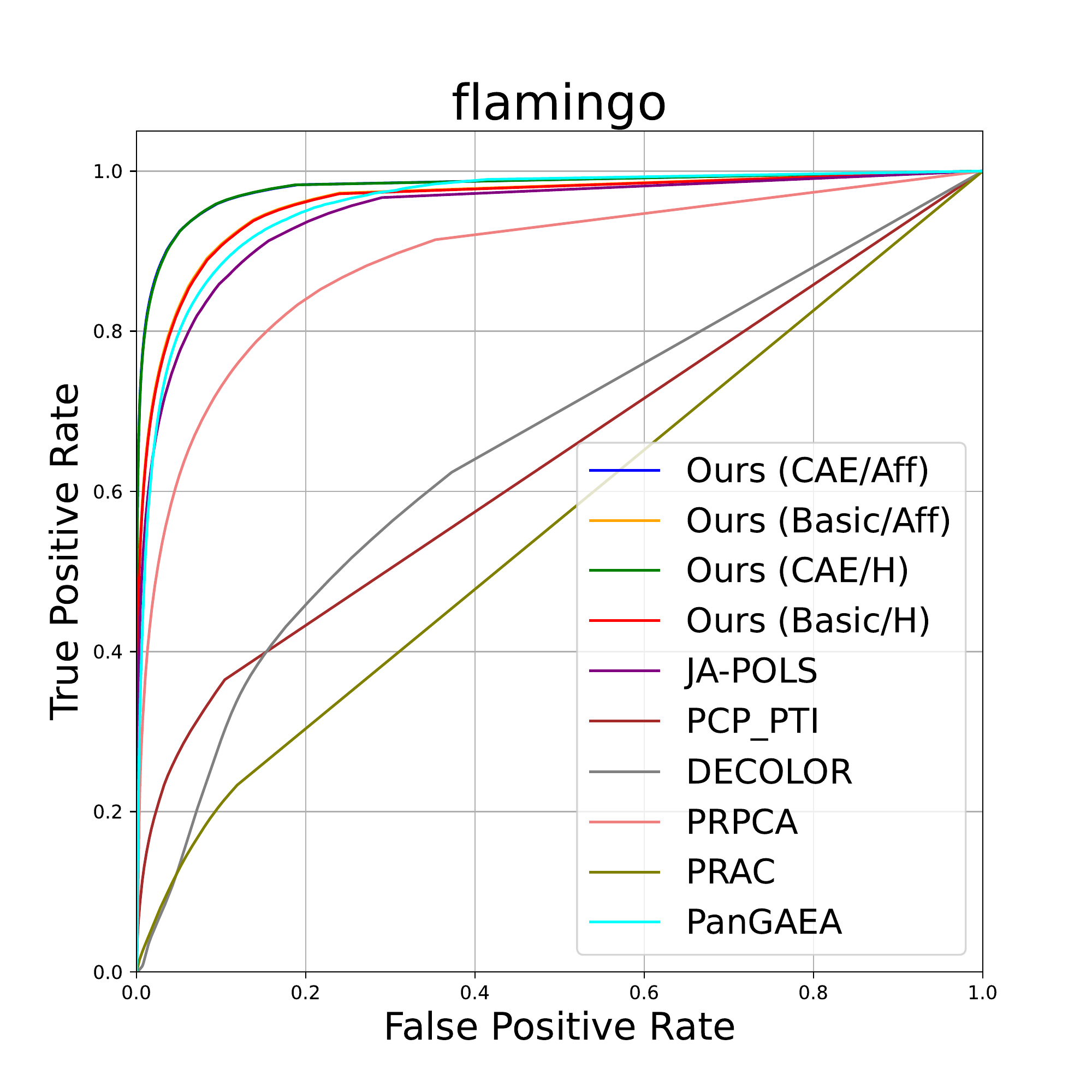}
    \includegraphics[width=\myW,trim={8mm 8mm 10mm 10mm},clip]{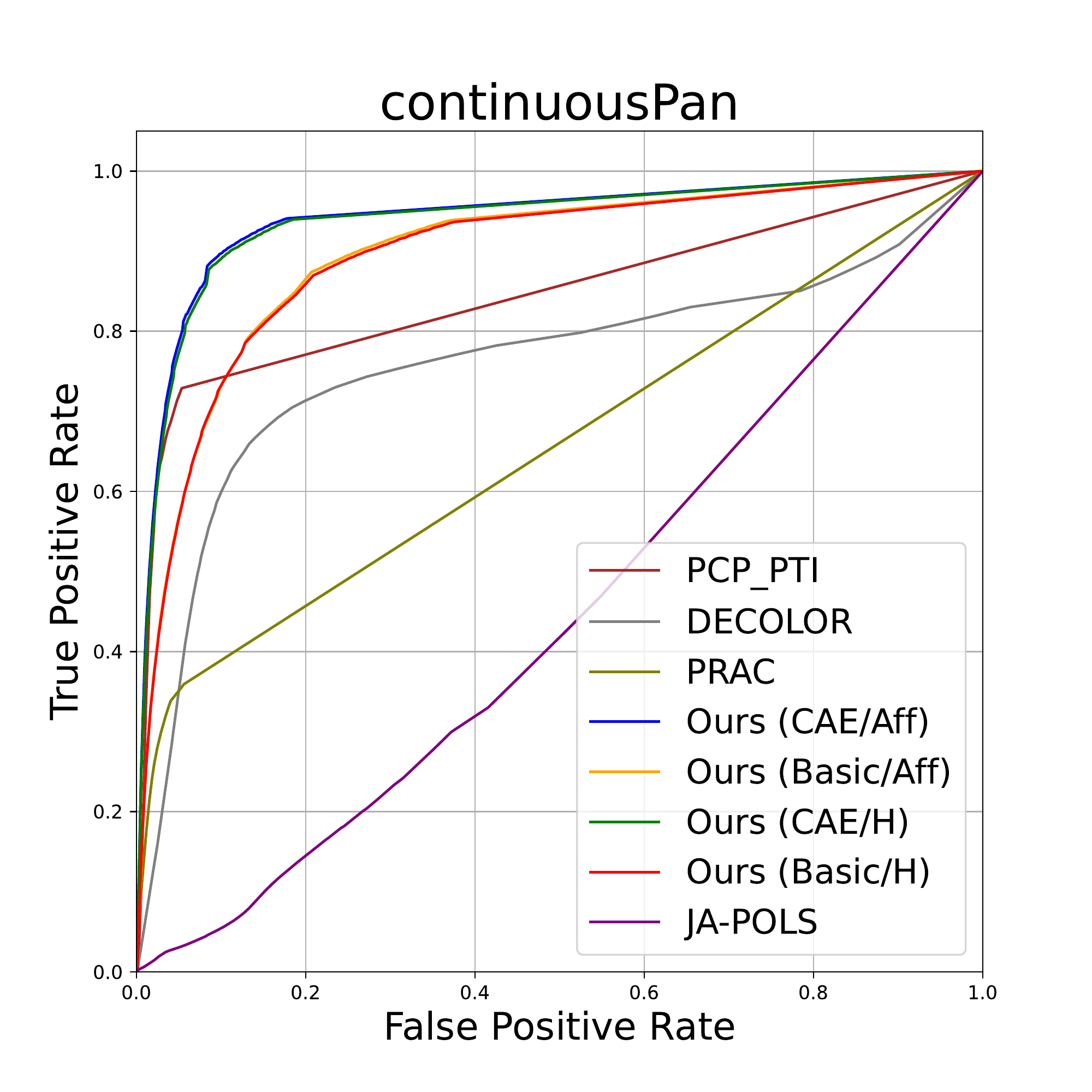}
    \includegraphics[width=\myW,trim={8mm 8mm 10mm 10mm},clip]{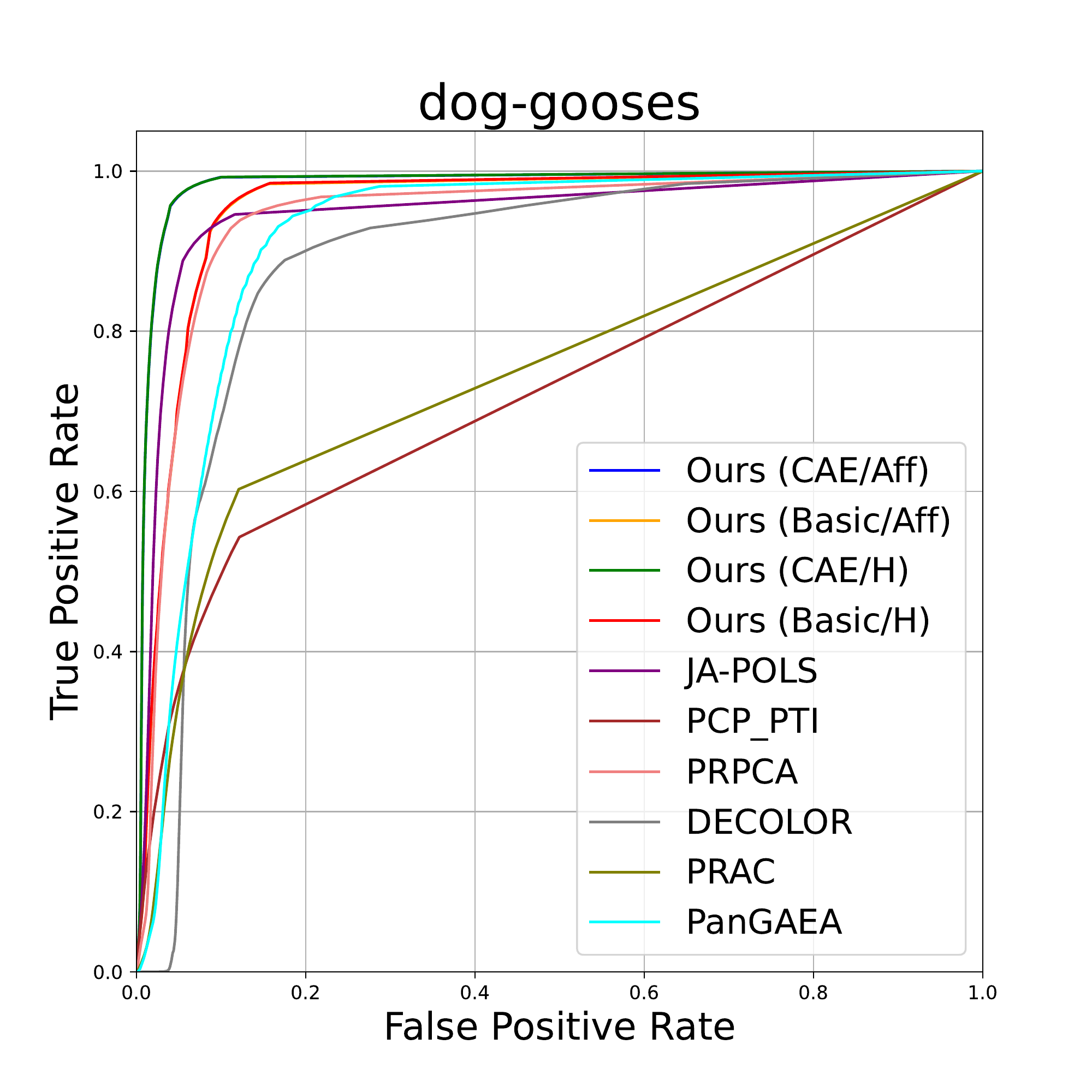}
    \includegraphics[width=\myW,trim={8mm 8mm 10mm 10mm},clip]{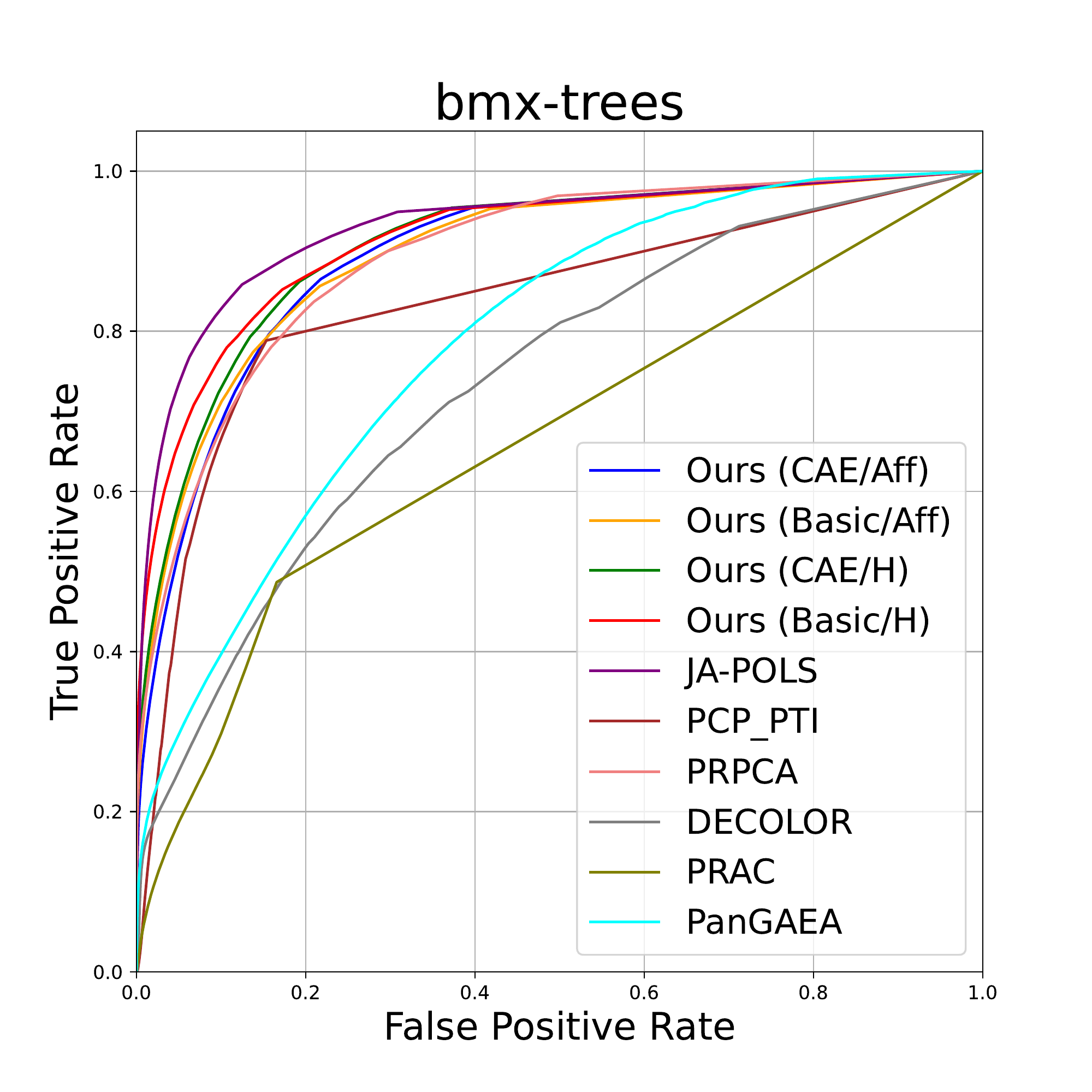}
    \includegraphics[width=\myW,trim={8mm 8mm 10mm 10mm},clip]{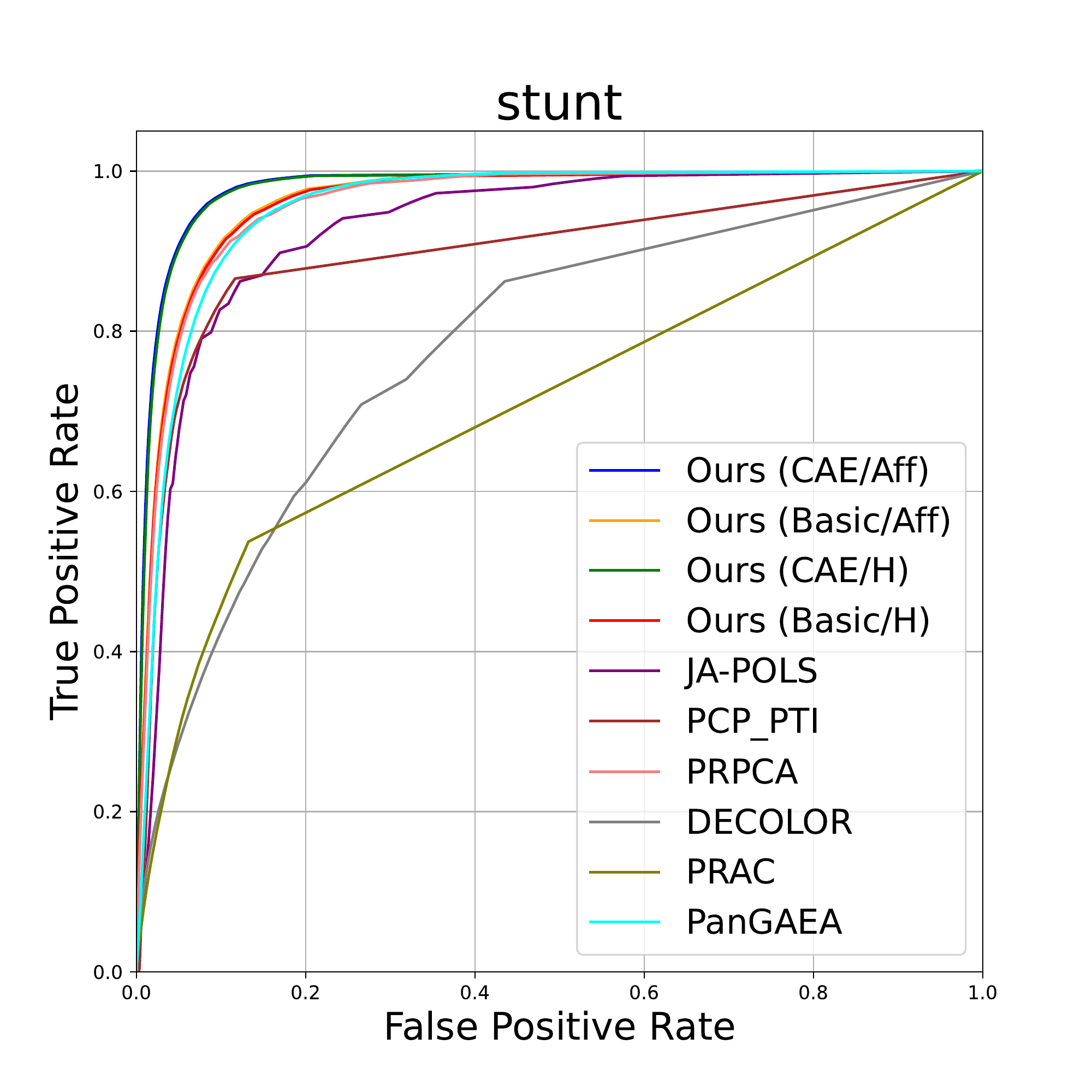}
    \includegraphics[width=\myW,trim={8mm 8mm 10mm 10mm},clip]{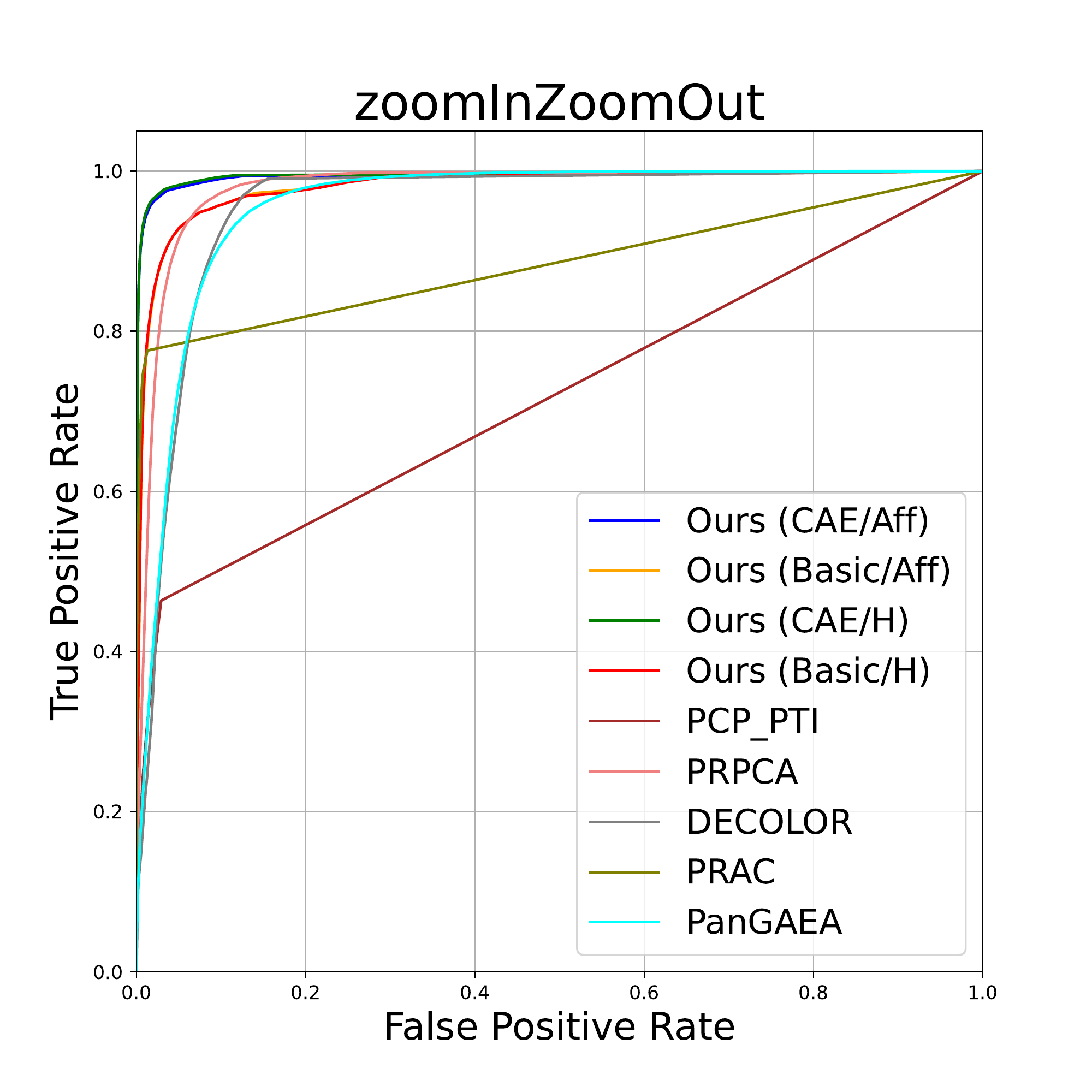}
    \includegraphics[width=\myW,trim={8mm 8mm 10mm 10mm},clip]{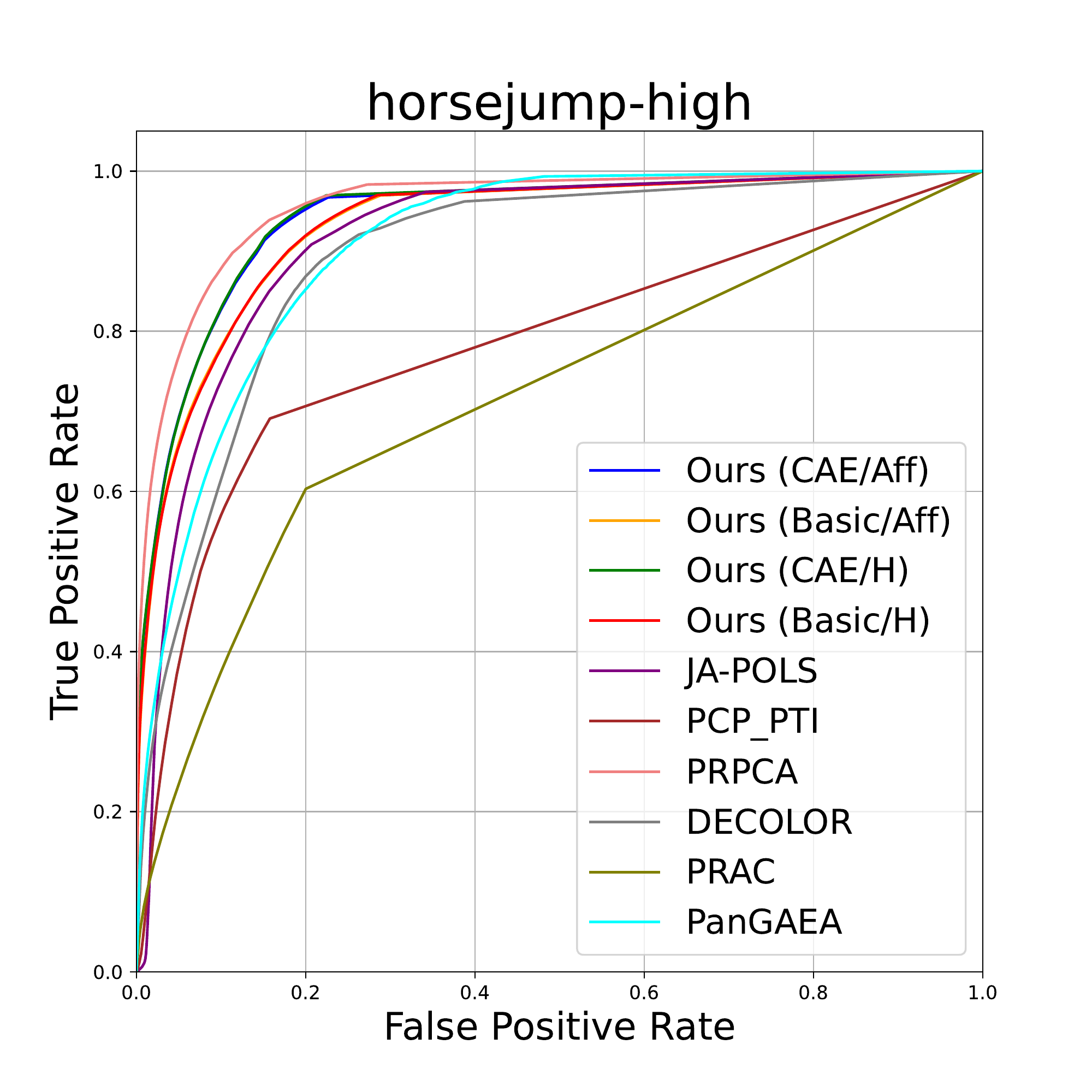}
    \includegraphics[width=\myW,trim={8mm 8mm 10mm 10mm},clip]{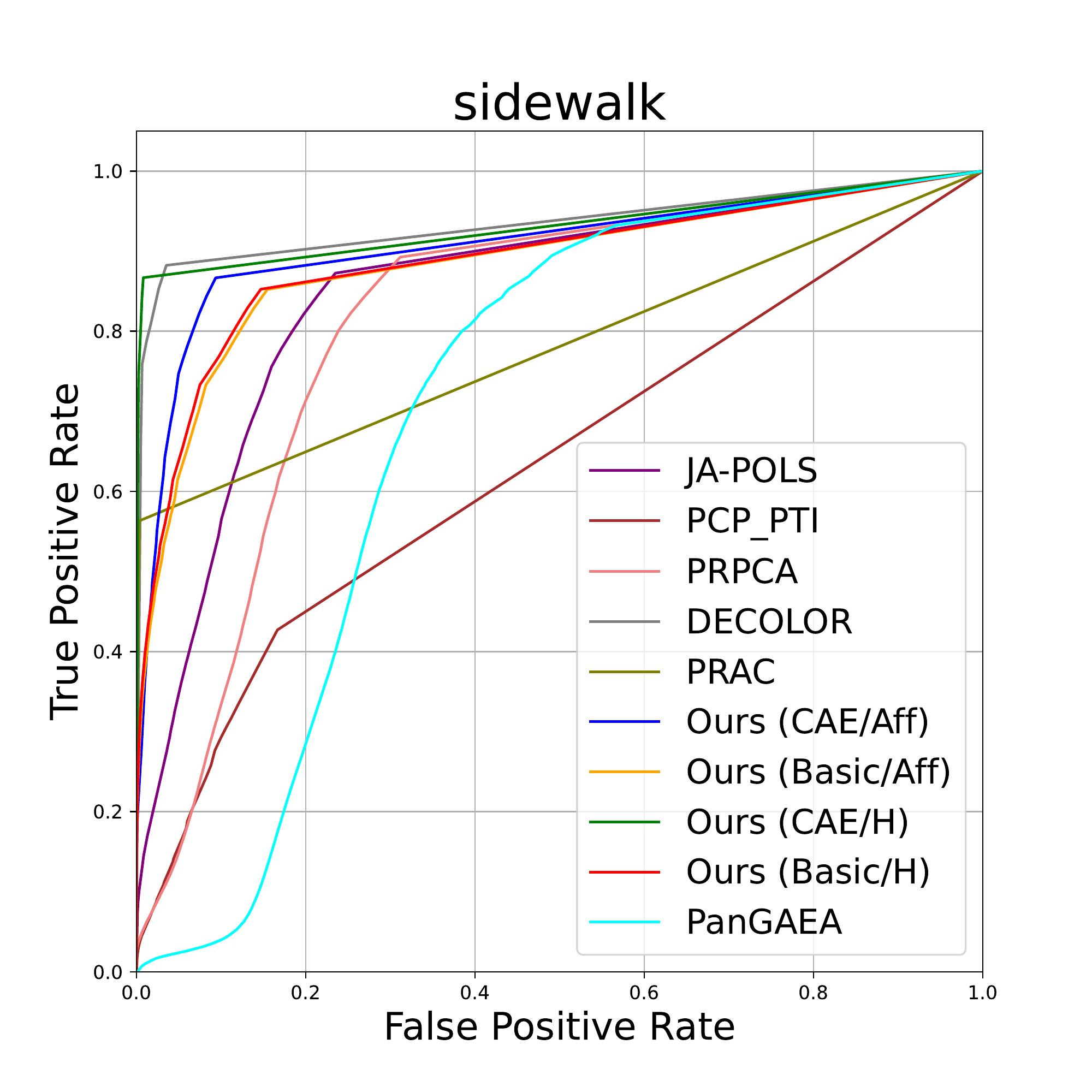}
    \includegraphics[width=\myW,trim={8mm 8mm 10mm 10mm},clip]{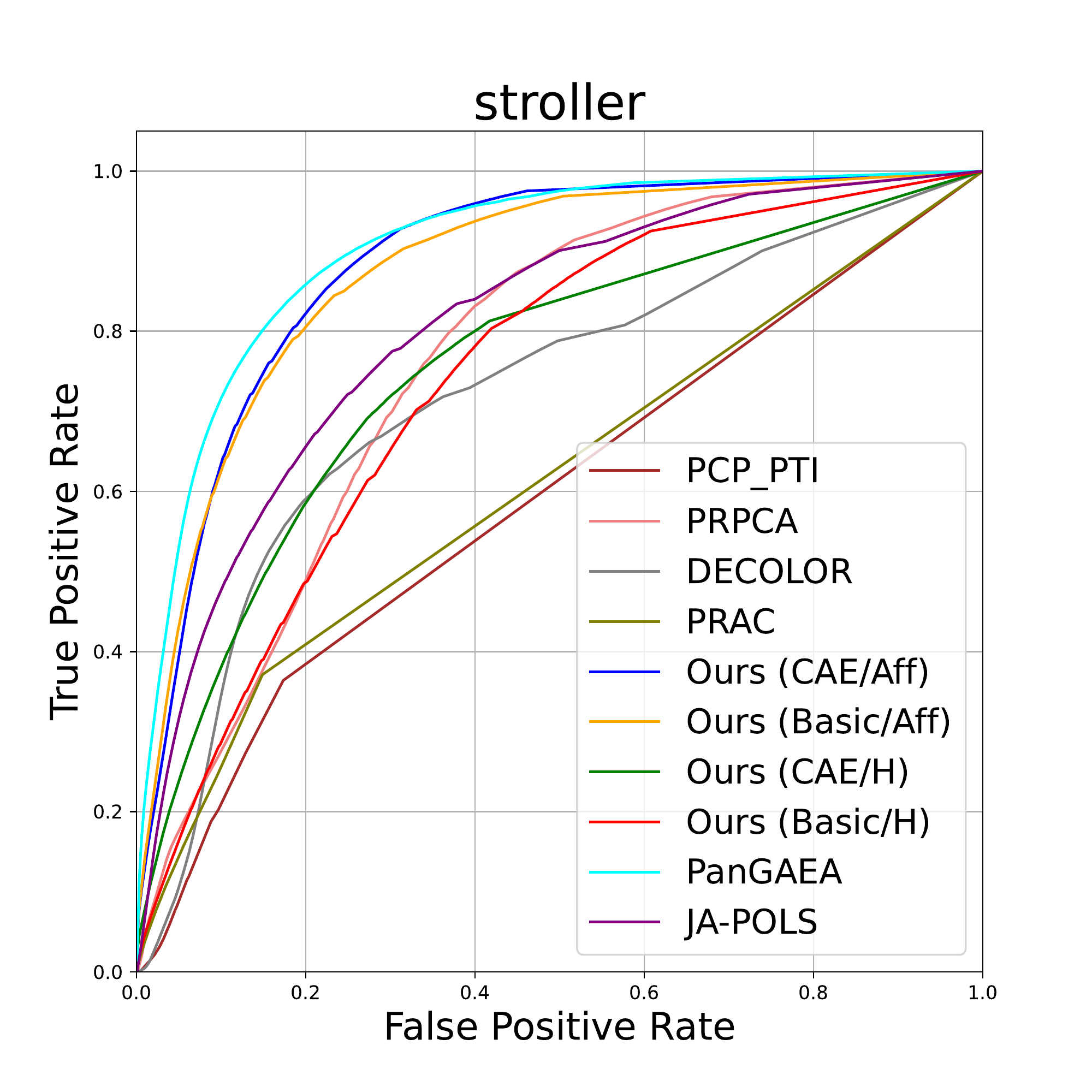}
    \includegraphics[width=\myW,trim={8mm 8mm 10mm 10mm},clip]{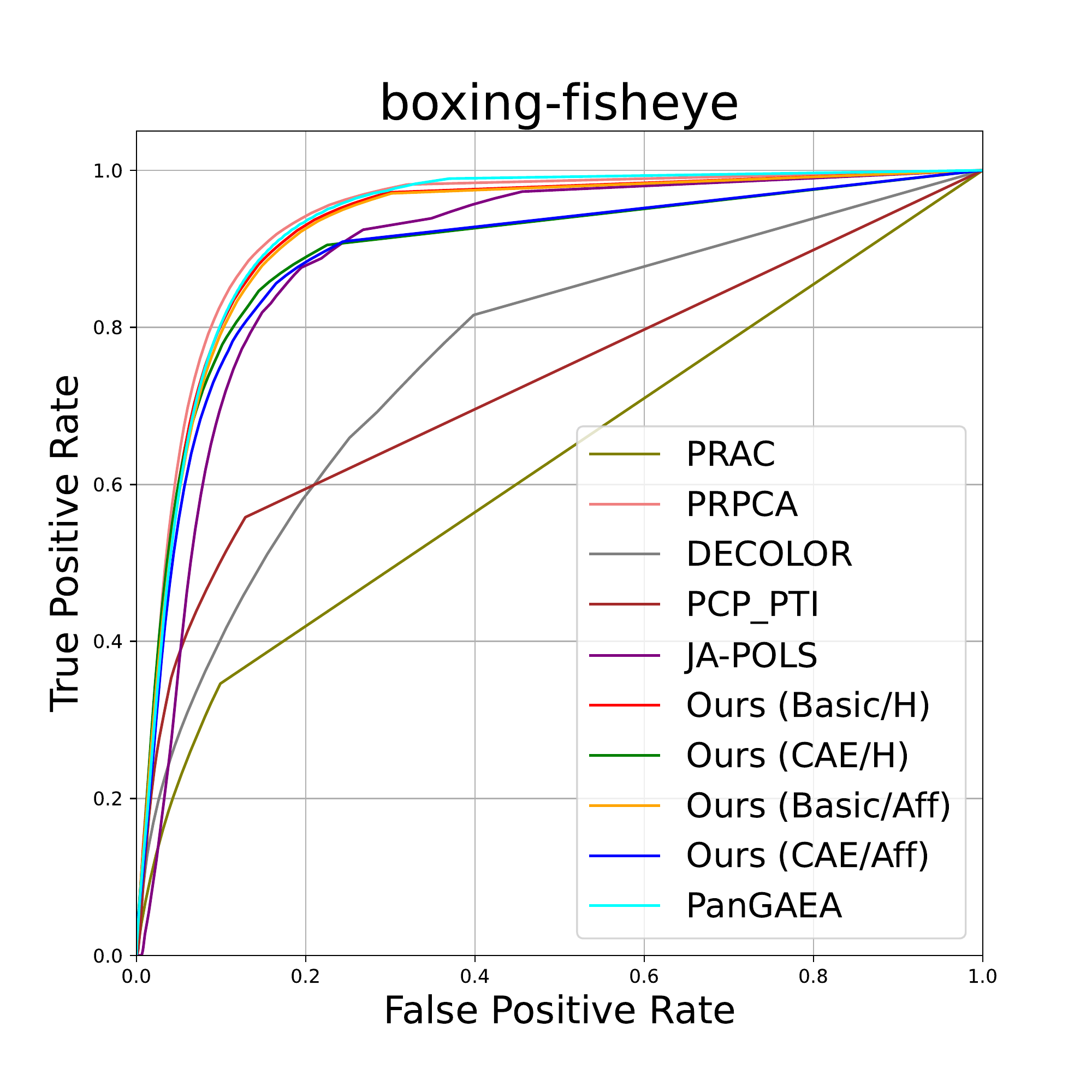}
    \includegraphics[width=\myW,trim={8mm 8mm 10mm 10mm},clip]{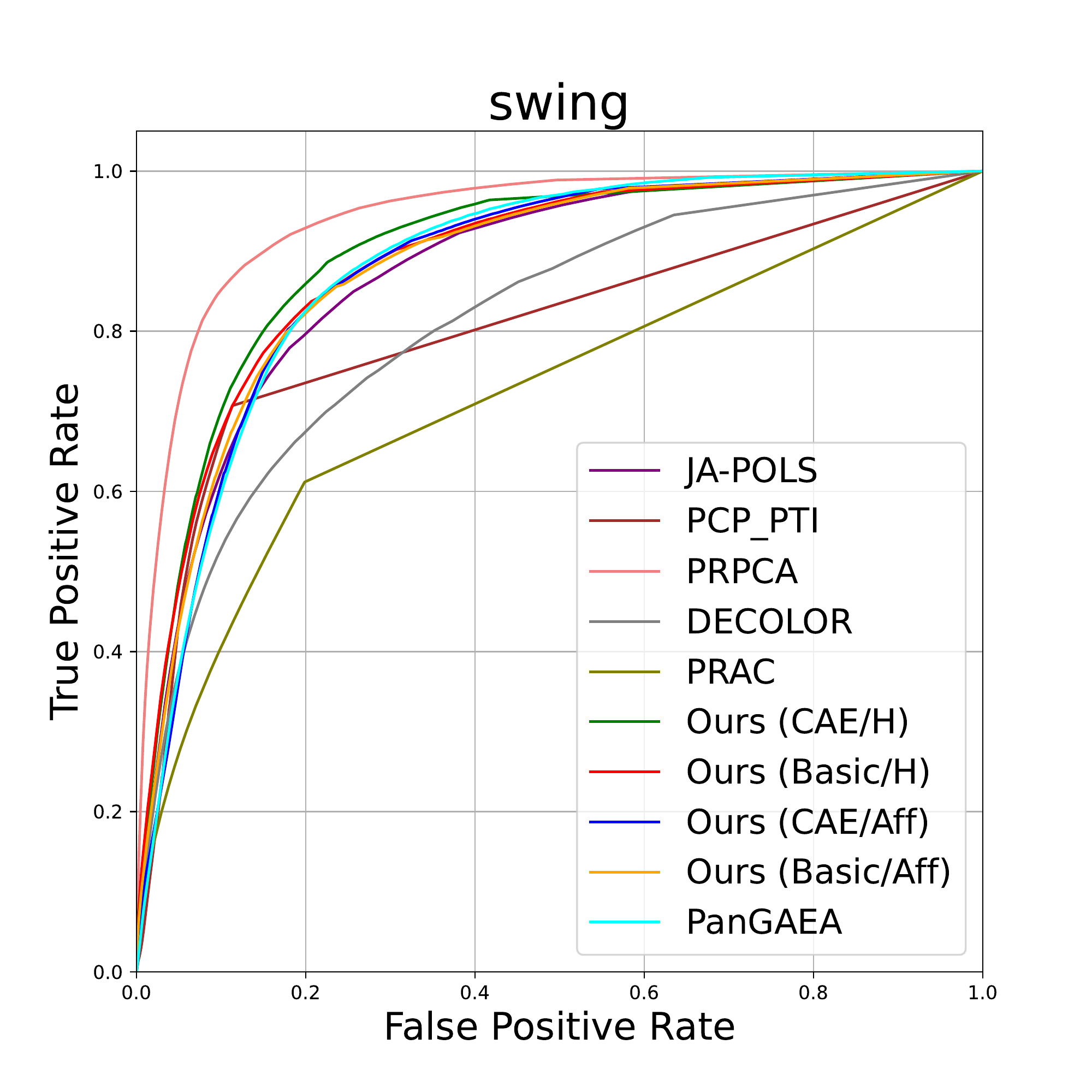}
    \includegraphics[width=\myW,trim={8mm 8mm 10mm 10mm},clip]{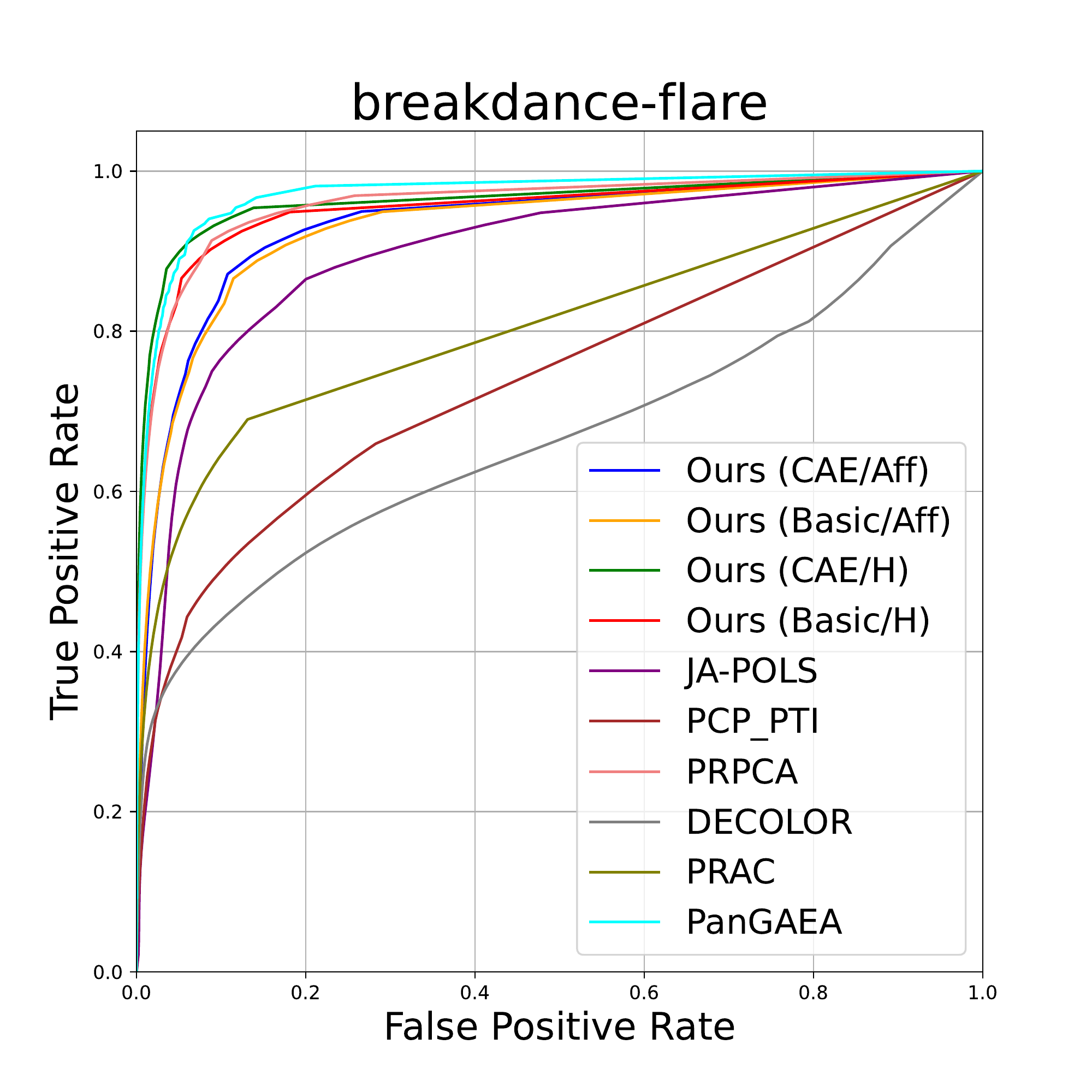}
    \caption{A comparison of ROC curves of different methods on various videos. Note that sometimes 
    some curves coincide (\eg, our CAE/Aff and CAE/Hom on the flamingo video). }
    \label{Fig:ROC}
\end{figure}

\section{Predicting Background to Previously-unseen Misaligned Frames}
In the paper, our comparison of DeepMCBM to the other methods
focused on their own playground: \ie, given an input sequence of video frames, the task  was viewed as an optimization problem whose overarching 
goal is to estimate the background in those frames.
However, like JA-POLS~\cite{Chelly:CVPR:2020:JA-POLS} but unlike the other competitors, DeepMCBM, being based on (unsupervised) learning, also possesses
a generalization capability in the sense it can estimate the background in previously-unseen misaligned frames from the video it was trained on. 
In contrast, other methods (JA-POLS excluded), do not have a readily-available mechanism that enables them to receive such misaligned frames and predict 
their alignment and background estimation -- at least not without solving additional optimization steps. Note that in a static camera background model
this is a non-issue since, by definition, the frames (both train and test)
are always aligned; however, in the moving-camera case the situation is different due to the misalignment. 
Remark: 
Note that this generalization capability should not be confused
with an online-learning setting -- which some competitors aim for -- where
each time the next consecutive frame arrives their model is being updated
(using further optimization).

\begin{figure}[ht]
    \newcommand{\myW}{0.29\linewidth}
        \centering
        \includegraphics[width=\myW,trim={8mm 8mm 10mm 10mm},clip]{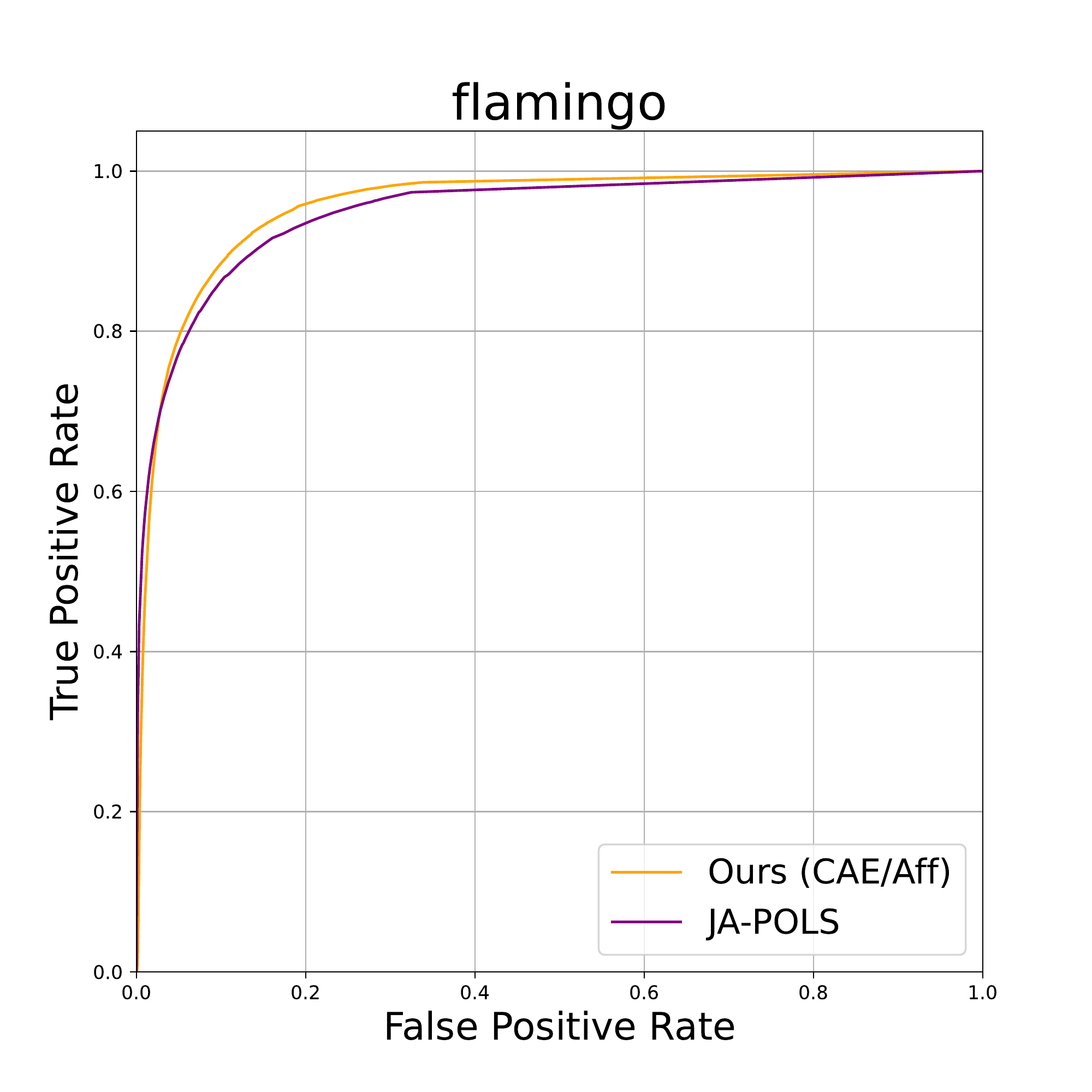}
        \includegraphics[width=\myW,trim={8mm 8mm 10mm 10mm},clip]{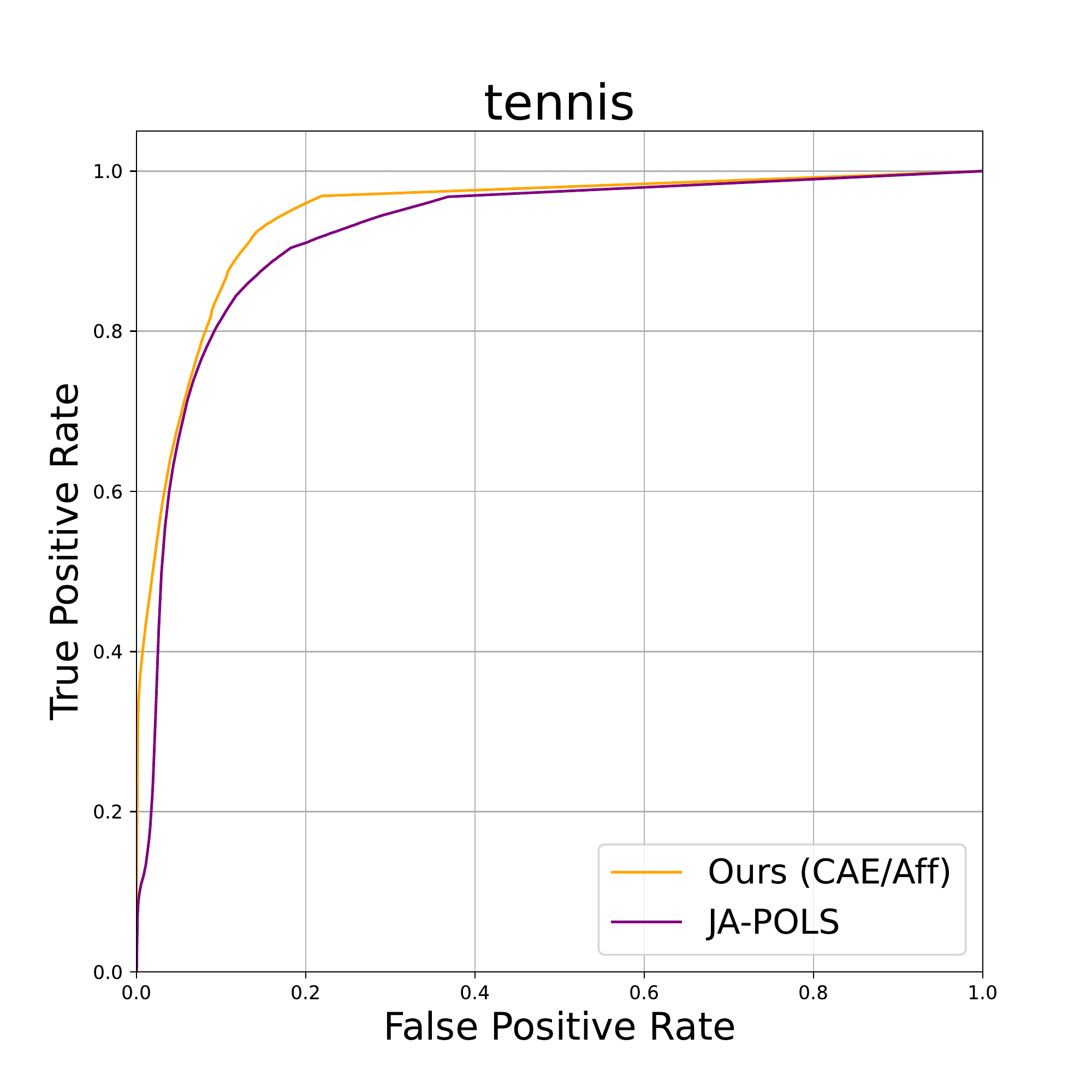}
        \caption{A typical comparison of ROC curves, between the proposed method and JA-POLS~\cite{Chelly:CVPR:2020:JA-POLS}, of the performance on test sets.         
        }
        \label{Fig:ROC_train_test}
    \end{figure}

To showcase the generalization capability, 
in each of the ``tennis'' and ``flamingo'' sequences (from~\cite{perazzi:2016:benchmark_davis}) we partitioned the video sequence into train and test sets.
We did it by letting the test set (in each of the two videos) consist of 10\%
of the frames (chosen at random from the original sequence) while letting the train test consist of the remaining 90\% of the frames.
In each video we let both DeepMCBM and its competitor, JA-POLS, train only on the train set. Next, we evaluated the models, using their prediction functionalities,
on the test sets.
\autoref{Fig:ROC_train_test} and 
\autoref{Table:AUC_train_test} summarize the results in terms of the corresponding ROC curve and AUC scores, respectively, and show that DeepMCBM outperforms
JA-POLS in this type of evaluation as well.  
Note that here we intentionally used our CAE/Aff variant (and not CAE/Hom)
to highlight the fact that even when using the same transformation type as JA-POLS (\ie, affine), DeepMCBM beats the latter.

\begin{table}[htbp]
    \centering
    \resizebox{0.5\textwidth}{!}{%
    \begin{tabular}{lcc}
    Sequence & DeepMCBM & JA-POLS \cite{Chelly:CVPR:2020:JA-POLS} \\ \hline
    \textbf{tennis test}        & 0.944 & 0.923  \\
    \textbf{tennis train}       & 0.965 & 0.936  \\
    \textbf{flamingo test}      & 0.957 & 0.949  \\
    \textbf{flamingo train}     & 0.980 & 0.949  \\
    \bottomrule     
    \end{tabular}%
  }
  \caption{AUC scores for test and train sets. DeepMCBM in its CAE+Affine version}     \label{Table:AUC_train_test}
  \end{table}
  
\section{The Robust Error Functions} 
We have used the  smoothed $\ellOne$ loss (which is closely-related
to Huber's function~\cite{Black:IJCV:1996:Robust})
for $\rho_\mathrm{JA}$
and Geman-McClure's error function~\cite{GemanMcClure:BISI:1987} for $\rho_\mathrm{recon}$:
\begin{align}
 \rho_\mathrm{JA}(\varepsilon;\beta) = \begin{cases}
    0.5\varepsilon^2/\beta & \quad\text{if}\quad|\varepsilon|\le\beta\\|\varepsilon|-0.5\beta&\quad\text{otherwise}\end{cases}
    & & \rho_\mathrm{recon}(\varepsilon;s) = 
    \frac{\varepsilon^2}{\varepsilon^2+s^2}\, 
\end{align}
where in all of our experiments we used $\beta= 0.35$ and $s=0.05$. 

\section{Evaluation Details}\label{Sec:Eval}
In this section we explain how the ROC curves and AUC scores were obtained.

Given an input sequence of $N$ frames, $({f^n})_{n=1}^N$,
\textbf{ground-truth annotations} of foreground pixels, $(a^n)_{n=1}^N$,
 and a background estimation sequence, $({\widehat{f}^n})_{n=1}^N$, 
we compute the pixelwise squared error $(E^n)_{n=1}^N$, where $E^n_\bx=\frac{1}{C}\sum_{c=1}^C(f^n_{\bx,c}-\widehat{f}^n_{\bx,c})^2$
is the error in pixel $\bx\in\Omega$ (recall that $\Omega$, a rectangle of height $h$ and width $w$, was defined in the paper as the common domain of the original input frames), averaged across $C$ channels.
To have a unified comparison measure applicable in all videos and for all methods, we scale to the $[0,1]$ interval where the scaling is done (for the method under consideration) \wrt the entire video sequence. That is, we define the scaled error
of that method in the specific video as
\begin{align}
    \widetilde{E}^n_{\bx} = \frac{E^n_{\bx}-\min_{n,\bx'}(E^n_{\bx'})}{\max_{n,\bx'}(E^n_{\bx'})-\min_{n,\bx'}(E^n_{\bx'})}\, .
\end{align}
Next, for each threshold value $\alpha$ (from a discrete set of evenly-spaced points between 0 and 1) we computed the True Positive Rate (TPR) and False Positive Rate (FPR): 
\begin{align}
    \mathrm{TPR} = \frac{1}{N\cdot h\cdot w} \sum_{n=1}^N\sum_{\bx\in\Omega}\mathbbm{1}(a_\bx = 1 \land \widetilde{E}^n_\bx \ge \alpha) \, ; \\
    \mathrm{FPR} = \frac{1}{N\cdot h\cdot w} \sum_{n=1}^N\sum_{\bx\in\Omega}\mathbbm{1}(a_\bx = 0 \land \widetilde{E}^n_\bx \ge \alpha) \, .
\end{align}
Computing the TPR and FPR for multiple threshold values lets us obtain the Receiver Operating 
Characteristic (ROC) curve for this (method,sequence) pair. 
\autoref{Fig:ROC} shows the resulted ROC curves while Area Under the (ROC) Curve (AUC) scores are summarized in Table 1 in the paper.

\section{Architecture and Training Details}
DeepMCBM's pipeline consists of two main parts:
\begin{enumerate}
    \item a Spatial Transformer Net (STN);
    \item a Conditional Autoencoder (CAE). 
    \end{enumerate}
    First, an STN is trained to learn an input-dependent function, that predicts the transformation parameter vector $\btheta$. 
    These predictions are used to create the panoramic central moments as well as, 
    later, to unwarp those moments back towards the input image.
    Once the STN training process has converged, the STN module is frozen
    and the CAE is trained to learn a background model in the input domain,
    where the conditioning is on the aforementioned unwapred moments. 
    Below we describe the architecture and training details of the STN and CAE
    modules. 
\subsection{The STN}
The STN we used consisted of a backbone and tailored regression heads for each available transformation type (in our case, either affine of homographies).
The backbone we used was ResNet18~\cite{Kaming:CVPR:2016:ResNet} whose output  size was $1000$. 
Each regression head consisted of two dense layers  of size 
$1000 \times 32$ and $32 \times d$, where $d$ is the dimension of 
the transformation parameters vector $\btheta$ ($d=6$ for affine transformations 
and $d=8$ for homographies). In both the regressor heads 
we used a ReLU~\cite{Agarap:2018:ReLU} activation function.
In all our experiments, we trained the backbone and the Affine head for the first $3000$ epochs.
In the variants that used homographies, 
we first did repeat the process above (that is, $3000$ epochs
for training the backbone and the affine head)
and then switched to train the homographic head
for additional $3000$ epochs 
(recall that our affine transformations are invertible and that such transformations are a particular case of homographies; 
thus, the results from the affine head provide a good initialization
for the homographic head). 
Thus, in total, the STN module was trained for about $6000$ epochs with a step learning rate scheduler,
decreasing the learning rate every $1000$ epochs with an initial learning rate of $0.005$.

\subsection{The CAE}
The second module in the DeepMCBM pipeline is a CAE, 
conditioned on the unwarped panoramic central moments. 
In our experiments, we conditioned the CAE on the first two such moments, 
but more generally, any  number of moments can be used as well.
Theoretically, using more moments implies that  
the distribution of the pixel stack is better captured. 

Our CAE was based on the (unconditional) Autoencoder from~\cite{Balle:ICLR:2017:AE}.
Using a similar architecture, we used 
$4$ channels in each hidden convolutional layer and $2$ channels for the last convolutional layer.
In between the convolutional encoder and decoder we used dense layers of size $256\times 4$ and $4 \times 256$ (thus, the code size was $4$) with a 
ReLU~\cite{Agarap:2018:ReLU} activation function. 
All convolutional layers used a $5\times 5$ kernel with stride size $2$.

Our conditioning is done in both the encoder and decoder parts. 
As the conditioning (the unwarped central moments) is in the same spatial dimensions as 
the input image, conditioning on the encoder size was done by a simple 
concatenation, along the channel dimension, of the input and the unwarped moments. 
This results in $(N_\mathrm{moments}+1)\cdot C$ input channels for the encoder
where $N_\mathrm{moments}$ 
is the number of the moments used ($N_\mathrm{moments}=2$ in our experiments) and $C$ is the number of channels of an input image ($C=3$ in the RGB case).     
To condition the decoder, we concatenate the unwaped moments to the (classic) decoder output 
and then use a small Convolutional Neural Net (CNN) to integrate the decoder's output and the conditioning. 
This last CNN is composed of 3 convolutional 
layers with 50 channels for the hidden layesr and $C$ channels for the output layer. 
All three layers use a $3\times 3$ kernel, and a ReLU~\cite{Agarap:2018:ReLU} activation function.

Using the robust reconstruction loss mentioned in the paper, we trained the CAE for $2000$ epochs with a learning rate scheduler and a step size 
of 500 epochs and initial learning rate of $0.001$.

\section{Using the Matrix Exponential within an STN}
Let $\bA$ be a $3\times 3$ real-valued matrix such that its last row
is all zeros:
\begin{align}
 \bA = \MATRIX{A_1 & A_2 & A_3 \\ A_4 & A_5 & A_6 \\ 0 & 0 & 0} \, .
\end{align}
Then, a known result
(which is also widely-used in computer vision; see, \eg,~\cite{Lin:CVPR:2009:Lie,Lin:CVPR:2010:Persistent})
from the theory of matrix groups
is
that 
\begin{align}
 \bT\triangleq \exp(\bA)
\end{align}
 (\ie, the matrix exponential of $\bA$)
 has the following form,
 \begin{align}
   \bT = \MATRIX{T_1 & T_2 & T_3 \\ T_4 & T_5 & T_6 \\ 0 & 0 & 1} \, 
 \end{align}
where, in addition, we have that $\det \bT>0$ (in particular, $\bT$ is invertible). 
Consequently, the matrix exponential provides a mapping from the unconstrained linear space $\RR^6$ into the Affine Group (namely, the space of invertible affine transformations -- in this case from $\Rtwo$ to $\Rtwo$). Taken together with the fact that the matrix exponential is a differentiable function, this means that an STN~\cite{Jaderberg:NIPS:2015:STN} can be easily set to produce only invertible transformations~\cite{Skafte:CVPR:2018:DDTN,Chelly:CVPR:2020:JA-POLS}.

\clearpage
%
%
\bibliographystyle{splncs04}
\bibliography{./refs}